\documentclass{bmvc2k}

\usepackage{caption}
\usepackage{algorithm}
\usepackage{algpseudocode}
\usepackage{lipsum}
\usepackage{multirow}
\usepackage{graphicx}
\usepackage{xcolor}
\usepackage{float}
\usepackage{lipsum}  
\usepackage{amssymb}

\title{Towards Debiasing Frame Length Bias in Text-Video Retrieval via Causal Intervention}

\addauthor{Burak Satar}{burak001@e.ntu.edu.sg}{1,2}
\addauthor{Hongyuan Zhu}{zhuh@i2r.a-star.edu.sg}{1}
\addauthor{Hanwang Zhang}{hanwangzhang@ntu.edu.sg}{2}
\addauthor{Joo Hwee Lim}{joohwee@i2r.a-star.edu.sg}{1,2}

\addinstitution{
 Institute for Infocomm Research\\
 A*STAR\\
 Singapore
}
\addinstitution{
 School of Computer Science and Engineering \\ 
 Nanyang Technological University \\
 Singapore
}

\runninghead{Satar et al.}{Debiasing Frame Length Bias in Text-Video Retrieval}

\def\etal{\emph{et al}\bmvaOneDot}

\begin{document}

\maketitle

\begin{abstract}

Many studies focus on improving pretraining or developing new backbones in text-video retrieval. However, existing methods may suffer from the learning and inference bias issue, as recent research suggests in other text-video-related tasks. For instance, spatial appearance features on action recognition or temporal object co-occurrences on video scene graph generation could induce spurious correlations. In this work, we present a unique and systematic study of a temporal bias due to frame length discrepancy between training and test sets of trimmed video clips, which is the first such attempt for a text-video retrieval task, to the best of our knowledge. We first hypothesise and verify the bias on how it would affect the model illustrated with a baseline study. Then, we propose a causal debiasing approach and perform extensive experiments and ablation studies on the Epic-Kitchens-100, YouCook2, and MSR-VTT datasets. Our model overpasses the baseline and SOTA on nDCG, a semantic-relevancy-focused evaluation metric which proves the bias is mitigated, as well as on the other conventional metrics.\footnote{https://buraksatar.github.io/FrameLengthBias/}

\end{abstract}

\section{Introduction}
\label{sec:intro}

In text-video retrieval, nowadays, the state-of-the-art models \cite{Liu_Chen_Huang_Chen_Wang_Pan_Wang_2022, Gorti_Vouitsis_Ma_Golestan_Volkovs_Garg_Yu_2022, 2022_centerclip, satar2022rome, satar2021semantic, satar2022exploiting} can achieve promising performance on famous benchmarks \cite{Xu_Mei_Yao_Rui_2016, Price_Vondrick_Damen_2022, Zhou_Xu_Corso_2017}. However, recent studies \cite{Yoon_debias_vcmr_2022, yang2021deconfounded, modality_bias_sigir_2022, satar2023overview} demonstrate that many existing visual-text models are overly affected by superficial correlations. For instance, some works \cite{single_frame_bias_2022, broome_bias_action_reg_wacv_2023, Hara_2021_CVPR_action_bias} address the static appearance bias for action recognition. While \cite{Chadha2021iPerceive} focuses on object co-occurrences that bring spurious correlations specifically in the spatial domain, \cite{Nan_2021_CVPR, Xu_meta_spatio_2022} examine the same topic in the temporal domain. Some other works reveal the correlation between the start-end time of the actions and the actions themselves in untrimmed videos on video moment retrieval \cite{Yoon_debias_vcmr_2022, yang2021deconfounded} and temporal sentence grounding \cite{lu_acm_sentence_grounding_2022, bias_sentence_ground_2021} tasks. 
Unlike these studies, we focus on a temporal bias that has yet to be addressed in text-video-related tasks. Frame length discrepancy between training and test sets of trimmed video clips causes non-relevant retrieved items. 

\begin{figure}
\begin{tabular}{ccc}
\bmvaHangBox{\fbox{\includegraphics[width=2.62cm]{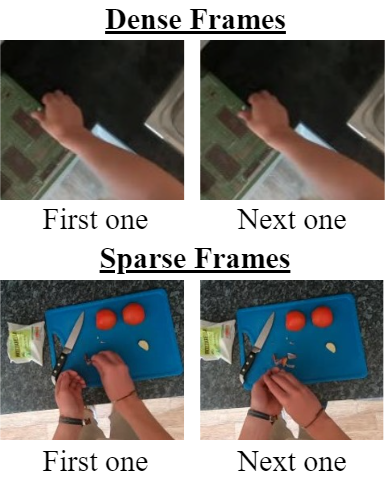}}}&
\bmvaHangBox{\fbox{\includegraphics[width=8.9cm]{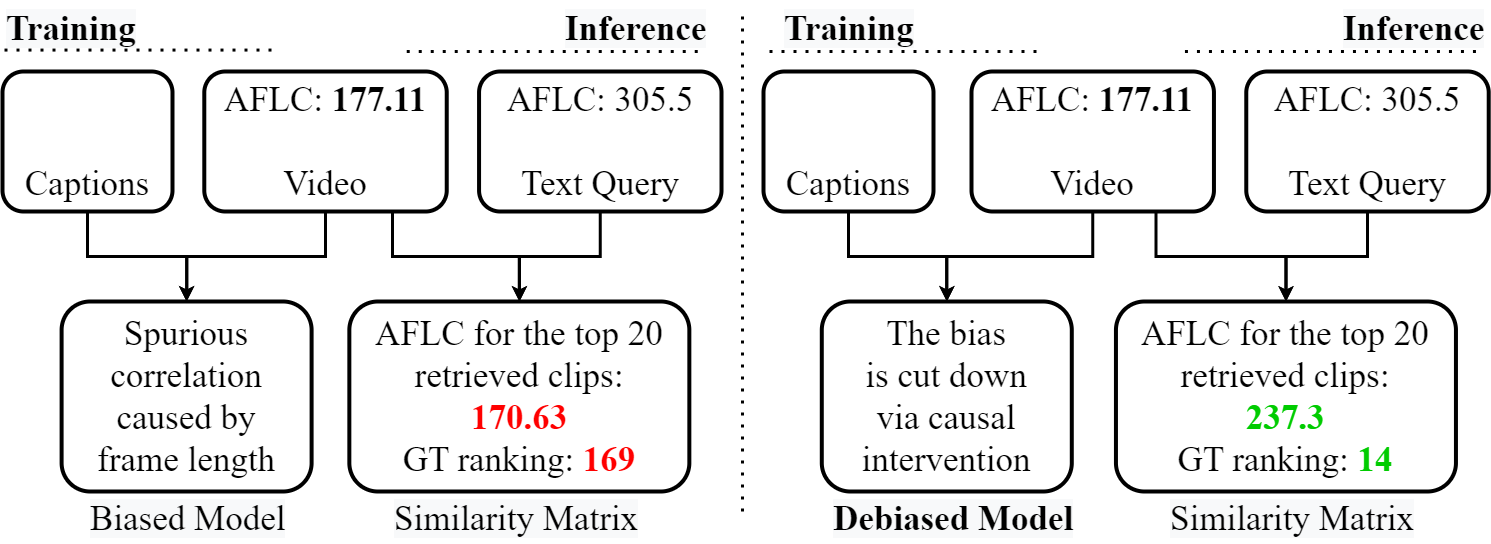}}}\\
(a)An illustration. %
& (b) Caption of the class: \textit{'pick up rubbish'}.
\end{tabular}
\vspace{0.15cm}
\caption{a) Motion semantics may differ between long and short video clips when the frames are uniformly sampled. If it is unbalanced between training/test sets, this may introduce frame length bias. b) AFLC denotes the average frame length of a class, meaning <verb, noun> pairs (classes) which affect the retrieved clips. We propose a novel causal intervention method to remove this spurious correlation. }
\label{fig:teaser}
\end{figure}

\begin{figure}[!htb]
\begin{tabular}{cc}
\bmvaHangBox{\fbox{\includegraphics[width=5.8cm]{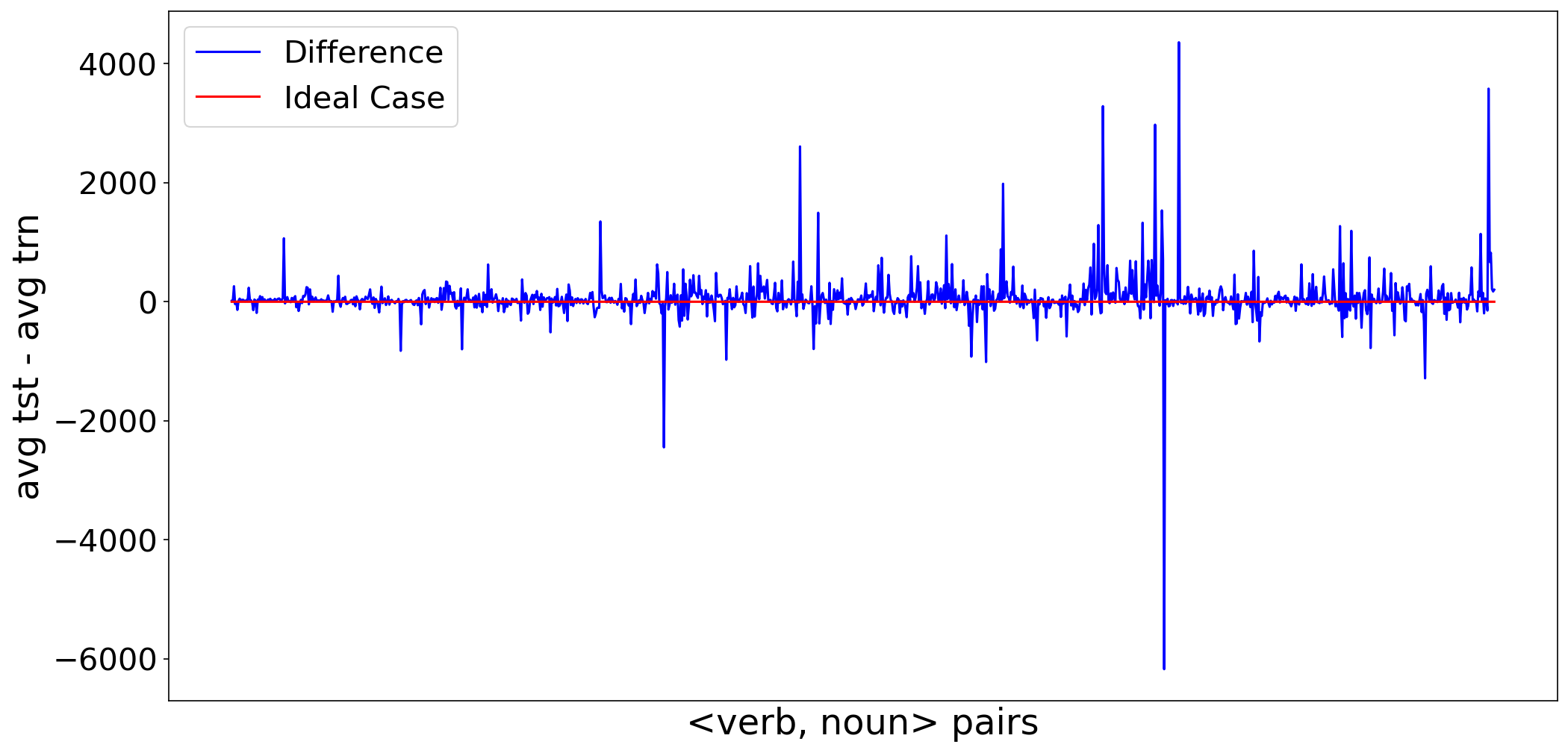}}}&
\bmvaHangBox{\fbox{\includegraphics[width=5.8cm]{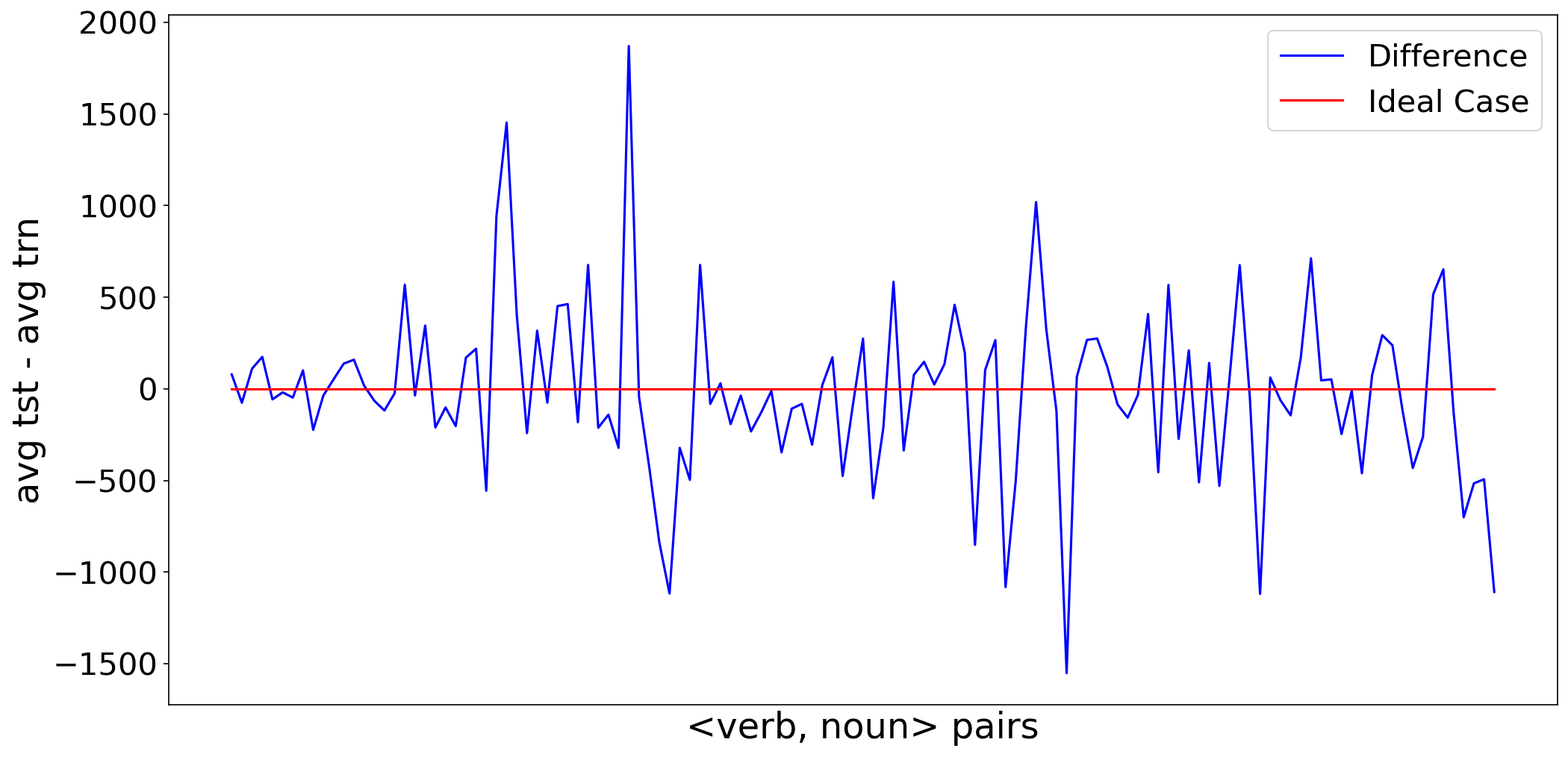}}}\\
(a) Epic-Kitchens-100 & (b) YouCook2
\end{tabular}
\vspace{0.15cm}
\caption{The figures show the discrepancy among all the <verb, noun> pairs (classes) in each dataset, which is calculated by the average frame length difference between the training and test sets. The number of pairs in the X-axis is 1,144 and 125, respectively. See the Supplementary Material for more details regarding verifying the bias in the three datasets.}
\label{fig:verify-bias}
\end{figure}

For example, in the case of text-to-video retrieval, as shown in Figure \ref{fig:teaser}, the top twenty retrieved clips' average frame length is similar to the training class's average frame length, stating that some irrelevant clips are retrieved just because of the bias coming from the discrepancy. We refer to 'class' as a joint combination of 'verb class' and 'noun class' by considering the verb and noun tokens together. Some classes can be semantically similar. For instance, 'take' and 'pick up' would be in the same verb class. Thus, we use the notion of class to calculate a matrix, measuring semantic relevancy among verbs and nouns. In addition, we utilise it to identify the biases in Figure \ref{fig:verify-bias}, showing the discrepancy. 
Only a recent work \cite{Zhan_Pei_Su_Wen_Wang_Mu_Zheng_Jiang_2022} closer to our approach attempts to mitigate video duration bias in watch-time prediction for video recommendation. However, the proposed model uses the video duration as textual input and does not consider any visual feature via any visual sampling method. They apply causal inference based on video duration while the discrepancy in the video duration is not considered, and it is followed by a pre-text task of watch-time prediction to increase the effect.

\noindent

To address overlooked frame length bias in text-video retrieval, we first apply baseline debiasing methods, which delete either the shortest or longest video clips in a class to reduce the discrepancy between train and test sets. However, the effect is limited. Then, we intervene in the causal graph to remove the frame length's unwanted impact by applying the backdoor adjustment principle \cite{judea_do_calculus}. Specifically, we divide the training data into splits regarding frame length; for each split, we learn a similarity matrix using the same text-video retrieval model. Then, we sum the similarity matrices. Note that we also consider the discrepancy within the splits regarding frame length to increase the debiasing effect. 
The contributions of this paper are threefold: \textbf{i)} To the best of our knowledge, we are the first to address a temporal bias in text-video retrieval tasks and also the first to address frame length bias in any text-video-related tasks. We verify the bias illustrated with various methods. \textbf{ii)} We propose a causal inference approach via backdoor adjustment to mitigate the frame length bias. \textbf{iii)} The experiments and ablation study verify the advantages of the proposed approach over the baseline and SOTA studies by evaluating retrieved clips semantically via Discounted Cumulative Gain (nDCG) as well as Recall and mAP.

\section{Related Work}

\textbf{Text-Video Retrieval.} In text-video retrieval, which aims to rank samples in a modality given another modality, deep learning-based approaches have emerged as promising techniques due to their ability to learn high-level features directly from the data. One popular method is to encode text and video features into a common space \cite{miech18learning, miech19howto100m, miech20endtoend}, where the similarity can be measured using various distance metrics. Another approach is to utilise the semantic relationships between text and video features \cite{chen_hgr_cvpr_2020, falcon2022learning, satar2021semantic}. For instance, Chen \etal \cite{chen_hgr_cvpr_2020} use semantic role labelling to capture the relationship between verbs and nouns in text and actions and objects in videos. Besides, Falcon \etal \cite{falcon2022learning} implements a positive and negative sampling strategy based on semantic similarities between verb and noun pairs. 
Recent models based on visual transformers have shown promising results with the help of pre-training on giant datasets \cite{miech19howto100m}. For example, Bain \etal \cite{Bain_Nagrani_Varol_Zisserman_2021} use raw video frames rather than extracted features and apply attention mechanisms for pre-training on various exocentric video datasets. On top of this work, Lin \etal \cite{egovlp} pre-train the modified model on an enormous egocentric dataset curated from Ego4D \cite{ego4d_dataset}. Nevertheless, further research is needed to address existing biases in the task.  

\textbf{Biases in Video-Language.}
Recent studies have highlighted the presence of biases in video-language tasks, which can affect the performance of models since the models can rely on spurious correlations in the data rather than genuine causal relationships. For instance, temporal \cite{Yoon_debias_vcmr_2022, Xu_meta_spatio_2022} and spatial \cite{broome_bias_action_reg_wacv_2023, Hara_2021_CVPR_action_bias, Chadha2021iPerceive, single_frame_bias_2022} biases may arise due to the nature of the data collection process, where certain activities or scenes may be over-represented or under-represented \cite{yang2021deconfounded, modality_bias_sigir_2022, bias_sentence_ground_2021}. In addition, certain words or phrases may be over-represented in the captions, leading to a bias towards those concepts \cite{Nan_2021_CVPR, lu_acm_sentence_grounding_2022}. However, various biases are overlooked, and it is crucial to understand the sources of these biases. In this respect, we address the frame length bias.

\section{Method}

\subsection{Base Model}

Given its state-of-the-art performance, we follow \textit{Chen et al.} \cite{chen_hgr_cvpr_2020} for the baseline work. The model contains two encoders, one for text and, one for video, and a text-video matching part. 

\textbf{Textual Encoding.} We disentangle the textual features hierarchically by utilising a pre-existing semantic role labelling tool \cite{shi2019simple} to comply with disentangled video features. For example, whereas a sentence could define global features, local features are represented by words that refer to actions and entities. We establish the connection between actions and entities as $r_{ij}$, where $i$ denotes action nodes and $j$ denotes entity nodes. Subsequently, the semantic role matrix $W_r$, which is designed to accommodate various semantic roles, is multiplied with initialised node embeddings $g^0_{i} = g_i \odot W_rr_{ij}$ such that  $g_i$ $\epsilon$ \{$g_e$, $g_a$, $g_o$\}. The one-hot vector $r_{ij}$ indicates the edge type from node $i$ to node $j$, while $\odot$ signifies element-wise multiplication. Then, a graph-attention network is employed to process adjacent nodes. $W_t$ matrix, which is utilised for all relationship varieties, exploits attended nodes, as shown in Eq. \ref{eq_text_enc}. When attention is applied to each node, the result is referred to as $\beta$. Once these formulas are applied, we obtain the textual representation for global and local features $c_i$ $\epsilon$ \{$c_e$, $c_a$, $c_o$\}; for sentence node, verbs, and words, respectively.

\begin{equation}
g^{l+1}_{i} = g^{l}_i + W^{l+1}_t \sum_{j\varepsilon  N_i}\beta_{ij} (g^l_j)
\label{eq_text_enc}
\end{equation}

\textbf{Video Encoding.} Disentangling videos into hierarchical features can be challenging, although it is comparatively simple to parse language queries into hierarchical features. To this end, we employ three distinct video embeddings that concentrate on various levels of video aspects. Given a video, denoted as V, represented as a sequence of frame-wise features $\sum_{i=1}^{M}f_i$ $\{f_1,...,f_M\}$, we apply different weights to generate embeddings for three different levels, which are then incorporated with a soft attention mechanism.

\begin{equation}
\begin{aligned}
v_{x,i} = \sum_{i=1}^{M} W^{v}_{x}f_i, \hspace{1cm} x\hspace{0.1cm}\epsilon \{e,a,o\}
\end{aligned}
\label{eq_video_enc}
\end{equation}

\textbf{Text-Video Matching.} The matching score is computed by averaging the cosine similarity with the video and textual embeddings. We use the contrastive ranking loss \cite{Chen_2020_CVPR} by attempting to have positive and negative pairs larger than a predetermined margin in training. Suppose $v$ and $c$ symbolise visual and textual representations; positive and negative pairs can be formulated as $(v_p,c_p)$ and $(v_p,c_n)$ / $(v_n,c_p)$, respectively. A pre-set margin named $Delta$ is used to determine contrastive loss.

\begin{equation}
s(V,C) = \sum^3_{i=1} \frac{<v_i, c_i>}{||v_i||_2 ||c_i||_2 }
\label{weights1}
\end{equation}

\begin{equation}
L(v_p,c_p) = [\Delta + s(v_p, c_n) - s(v_p,c_p)] + [\Delta + s(v_n, c_p) - s(v_p,c_p)]
\end{equation}

\subsection{Baseline Debiasing Method}

\begin{algorithm}[!htb]
\scriptsize
\caption{Delete the shortest clips. For each class in the common class set:}
\begin{algorithmic}[1]
\State $V$ $\gets$ videos in ascending order based on the frame length 
\State $x$ $\gets$ avg frame length of class for the training set \& $y$ $\gets$ avg frame length of class for the test set
\While{$y \geq x + \delta$}
\State Delete the first clip $v_0$ from the training set
\If{len(V) $\leq \alpha$} 
    \State break;
\EndIf
\EndWhile
\end{algorithmic}
\end{algorithm}

We can naively remove this bias by following two methods. In the first method, \textit{RmvOne}, we delete the shortest and longest class samples so that the training set's average frame length becomes similar to the test set for only one class. Note that the class notion refers to <verb, noun> pairs to group the captions semantically. However, this method does not affect the evaluation metrics, but only a few samples. Thus, another simple method, \textit{RmvAll}, can be suggested. We do the same as in \textit{RmvOne}, but considering all classes such that the high discrepancy will be reduced in the whole dataset to a pre-set margin $\delta$. We set the minimum number of video clips of a class in training as $\alpha$ so that there are enough samples for each class. It aims to reduce discrepancies between training and test sets for the same classes. Specifically, Algorithm \textcolor{red}{1} presents the way if the average test set is higher than the average train set and removes the shortest clips. The same logic applies when the situation is the opposite, which deletes the longest clips. 

\subsection{Method with Causal Intervention}

\begin{figure}[!htb]
    \centering
    \resizebox{8cm}{!}{
        \includegraphics{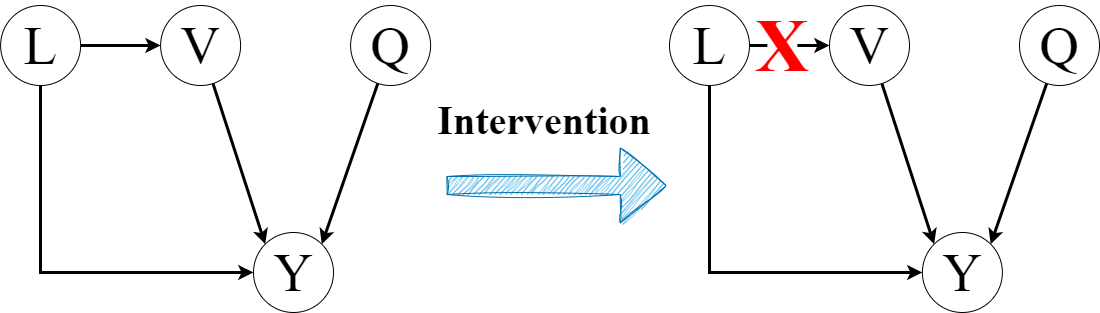}
    }
    \vspace{0.3cm}
    \caption{Structural causal model.}
    \label{fig:structural_causal_model}
\end{figure}

Many works use extracted features that are uniformly sampled in order to remove the effect of frame length. However, these features may still contain bias due to the sparsity or density of the sampled frames, as shown in Figure \ref{fig:teaser}. Thus, the model learns that action should be dense or sparse rather than motion semantics. The ideal case would be to have all the video clips at the same length. It is not just impractical but would also not reflect real-world applications. For example, while some actions take more time, others take less time intrinsically. Thus, while we need to keep this natural connection, we should remove the spurious correlation on video features that would occur because of the discrepancy in terms of frame length.
Figure \ref{fig:structural_causal_model} shows our structural causal model (SCM) to illustrate how our model works. V, Q, Y and L denote video representation, textual representation, text-video matching and frame length, respectively. The link from (V, Q) to Y is for capturing the similarity between the textual and visual features. The link from L to Y signifies the frame length effect on similarity, suggesting that while some actions can take less time, others would take more time. Moreover, the link from L to V implies that frame length would affect the video encoder such that various videos could be retrieved not because of their semantic similarity to the query but instead of their frame length. If this bias is not addressed,  densely sampled video features would be memorised in case the training set contains mostly shorter clips than the testing set. 

\begin{figure}[!htb]
    \centering
    \resizebox{11.9cm}{!}{
        \includegraphics{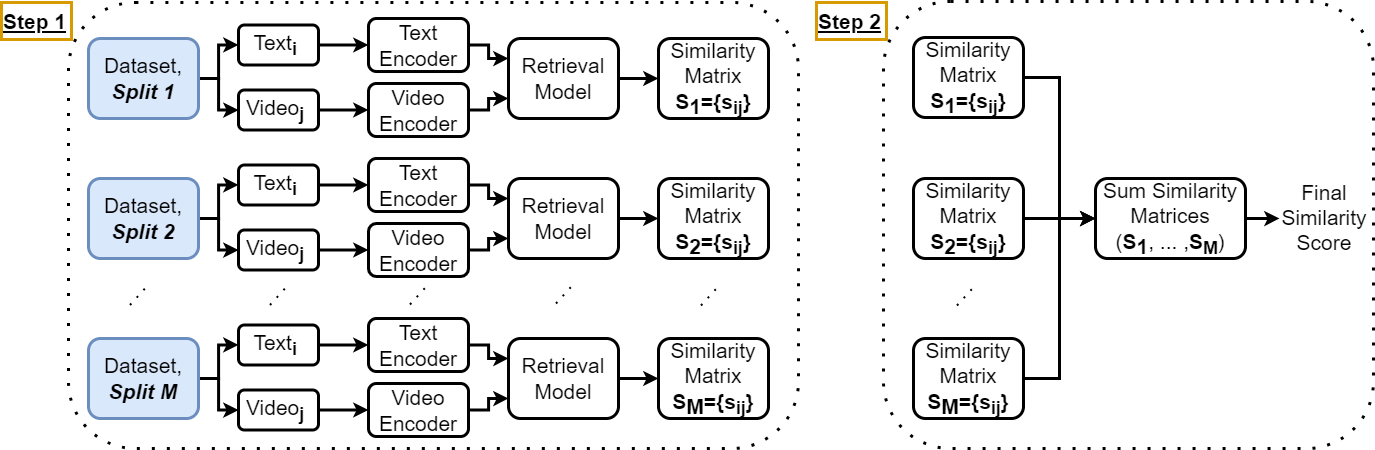}
    }
    \vspace{0.3cm}
    \caption{The implementation of the causal model for training. Similarity matrices are constructed using the same retrieval model on different splits that are arranged with a causal perspective to mitigate frame length bias. Then, they are summed up. No change is needed for the inference.}
    \label{fig:model_design}
\end{figure}

Figure \ref{fig:model_design} shows the implementation of two splits for a dataset based on the frame length. As a high-level idea, we follow the principle of backdoor adjustment to remove bias by splitting the dataset based on frame length. We formalise our causal method in Formula \ref{formula_causal} by using the law of iterated expectations. Note that L becomes independent via interventions. As shown in the last row of the formula, the final estimation can be created by individually estimating $P(L)$ and ${E}[Y|V,Q,L]$ and then combining those estimates. We divide the training samples into $M$ equal portions based on frame length to cut off the link, discretising the $P(L)$ distribution into separate components. These frame length groups are denoted by $\{L_k\}^{M}_{k=1}$. We estimate the deconfounded model via this approximation. Note that $f_k(v,q)$ is the similarity score for each frame length group $L_k$.

\begin{equation}
\begin{aligned}
    {E}[Y|do(V,Q)] &= \sum_{l}P(L=l|V,Q){E}[Y|V,Q,L=l] \\
    &= \sum_{l}P(L=l){E}[Y|V,Q,L=l] \\
    &\approx \sum_{k=1}^{M}(L_k){E}[Y|V,Q,L \in L_k] \\
    &\triangleq \sum_{k=1}^{M}(L_k)f_k\{V,Q\}
    \label{formula_causal}
\end{aligned}
\end{equation}

\section{Experiments}

\textbf{Datasets.} We use three datasets for our experiments. \textbf{i)} \textit{Epic-Kitchens-100 (EK-100)} \cite{Damen2022RESCALING}, a collection of unscripted egocentric action data gathered worldwide using wearable cameras. The annotated videos display diverse daily kitchen activities, accompanied by captions provided by human annotators that include at least one verb and one or more nouns. The dataset comprises 67,217 training and 9,668 test set pairs. \textbf{ii)} \textit{YouCook2} \cite{Zhou_Xu_Corso_2017, ZhXuCoAAAI18}, which is from cooking-related videos via third-person viewpoint collected from YouTube with 89 different recipes. The video clips are recorded from a third-person viewpoint within diverse kitchen settings. Imperative English sentences and temporal boundaries referencing the actions are used to label the video clips, and human annotators are used. There are 10,337 pairs in the training set and 3,492 pairs in the test set. \textbf{iii)} \textit{MSR-VTT dataset} \cite{xu2016msr-vtt} comprises 10,000 video clips with 20 descriptions for each video, a combination of human annotation and a commercial video search engine. The dataset offers several train/test splits, with one of the most popular ones being the 1k-A split, consisting of 9,000 clips for training and 1,000 clips for testing. The full split, which consists of 6,513 video clips for training, 2,990 video clips for testing, and 497 video clips for validation, is another often-used split.

\textbf{Implementation details.} We use the video features that TBN \cite{Kazakos_Nagrani_Zisserman_Damen_2019} has extracted for Epic-Kitchens-100. Each video clip included RGB, flow, and audio features. Note that we use the frame itself rather than using extracted features for replicating a sota work, EgoVLP. We utilise S3D features from \textit{Li et al.} \cite{li2021value} pretrained on HowTo100M\cite{miech19howto100m} for YouCook2. Since the test set is not made available to the public, we feed our model with the validation dataset for evaluation in accordance with other studies. For MSR-VTT, appearance level features of the ResNet-152 model provided by Chen \textit{et al.} \cite{Chen_2020_CVPR} are implemented. The epoch is chosen as 100 for all. $\Delta$ is determined as 0.2 by following the baseline model. $\delta$ and $\alpha$ are chosen as 10 and 60fps, respectively, for our baseline debiasing method. For SOTA methods RAN and RANP, negative and positive sampling thresholds are selected as 0.75 and 0.20, respectively. We report the best results out of three repetitions.

\textbf{Evaluation metrics.} We use the nDCG \cite{nDCG_formula} by considering non-binary similarity to show how the bias affects various retrieved video clips and how the causal model mitigates the bias. Given a caption query $q_i$ and a ranked list of video clips $X_r$, it \cite{wray2021semantic} is defined as $nDCG(q_i, X_r)=\frac{DCG(q_i, X_r)}{IDCG(q_i, X_r)}$. Then, DCG is calculated as $DCG(q_i, X_r)=\sum^{N_r}_{j=1}\frac{R(q_i, x_j)}{log_{2}(j+1)}$, and the ranking list only considers the first $N_r$ items, while $x_j$ is the $j$-th item in the list $X_r$. $IDCG$ is calculated via $nDCG$ and is the ideal case where $X_r$ is ordered by relevance. R, the relevancy matrix, is between 0 and 1 and represents the mean Intersection over Union (IoU) for the verb and noun classes. We follow \cite{Damen2022RESCALING} to define the R matrix as between a caption $q_i$ and a video $x_j$ by averaging the IoU of verb and noun classes. While $q_i^v$ refers to the collection of verb classes in the caption, $x_k^N$ denotes the set of noun classes in the video clip.

\begin{equation}
    R(q_i, x_j) = \frac{1}{2}\Bigg(\frac{|q_i^v \cap x_j^v |}{|q_i^V \cup x_j^v |} + \frac{|q_i^N \cap x_j^N |}{|q_i^N \cap x_j^N |}\Bigg)
    \label{relevancy}
\end{equation} 

By using the same logic, $nDCG$ is defined for a query video $x_i$ and a set of captions $C_r$. We follow the scripts provided by \cite{wray2021semantic} to create the relevancy matrices for the datasets. We utilise the mean average precision (mAP) and Recall (R@k) for a fair comparison.

\subsection{Results}

\textbf{Quantitative Results.} 
The first baseline debiasing method, \textit{RmvOne}, is impractical to repeat for all video classes, although it works for many examples. Table \ref{tab:result_epic_1} shows a result on the following baseline debiasing method, \textit{RmvAll}, for Epic-Kitchens-100. Specifically, we delete 2,392 clips from 164 classes, equivalent to 3.6\% of all the data, applying Algorithm \textcolor{red}{1}. It reaches marginally higher results on nDCG, even though it uses fewer data for training. Considering that we lose some information for many classes, such as diverse and complex visual cues, it is reasonable not to see a sharp increase by this naive method. We also compare it to the model that randomly deletes the same amount of video clips called \textit{RmvRand}, showing that knowing which clips to remove is essential. Although the ensemble approach overpasses the baseline, it is still lower than our method. Besides, its training takes three times more than ours; more importantly, its nDCG score is much lower than our approach, showing that the ensemble method does not address the bias as much as our causal model.

Tables \textcolor{red}{1}-\textcolor{red}{3} show the results of the causal method when M is chosen as 2. Specifically, the dataset is divided into two splits based on the frame length by considering the distribution of the dataset. Rather than having equal splits, we make one split that has more video clips than the other to have less discrepancy within the splits in terms of frame length. We choose the mean length of the test set as a threshold for splitting. Considering the baseline comparison, the average scores for nDCG increase by more than 2 points in each dataset. We see a similar trend for Recall and mAP metrics. A reasonable increase is observed when we apply our method to SOTA methods. Since these methods implement a specific scheme for positive and negative sampling, they force fewer pairs to match anchor samples when the dataset is split, which may limit the increase. Note that 'T2V' refers to text-to-video, and 'V2T' refers to video-to-text. Refer to the Supplementary Material to see the results of the causal method on the MSR-VTT's full split and more detail on baseline debiasing experiments.

\begin{table}[!htb]
\begin{center}
\parbox{.55\linewidth}{
\begin{center}
\resizebox{6.3cm}{!}{%
\begin{tabular}{|ccccccc|}
\hline
\multicolumn{1}{|c|}{\multirow{2}{*}{\textbf{Method}}}                                    & \multicolumn{3}{c|}{\textbf{nDCG}}                                                                                                                                                                                                            & \multicolumn{3}{c|}{\textbf{mAP}}                                                                                                                                                                                        \\ \cline{2-7} 
\multicolumn{1}{|c|}{}                                                                    & \multicolumn{1}{c|}{\textbf{V2T}}                                              & \multicolumn{1}{c|}{\textbf{T2V}}                                              & \multicolumn{1}{c|}{\textbf{AVG}}                                             & \multicolumn{1}{c|}{\textbf{V2T}}                                              & \multicolumn{1}{c|}{\textbf{T2V}}                                              & \textbf{AVG}                                             \\ \hline
\multicolumn{7}{|c|}{\textbf{Epic-Kitchens-100}}                                                                                                                                                                                                                                                                                                                                                                                                                                                                                                                          \\ \hline
\multicolumn{1}{|c|}{Baseline}                                                            & \multicolumn{1}{c|}{39.40}                                                     & \multicolumn{1}{c|}{38.91}                                                    & \multicolumn{1}{c|}{39.15}                                                    & \multicolumn{1}{c|}{40.47}                                                    & \multicolumn{1}{c|}{36.60}                                                     & 38.54                                                    \\ \hline
\multicolumn{1}{|c|}{\begin{tabular}[c]{@{}c@{}}Baseline +\\ RmvRand\end{tabular}}           & \multicolumn{1}{c|}{39.69}                                                    & \multicolumn{1}{c|}{38.42}                                                    & \multicolumn{1}{c|}{39.06}                                                    & \multicolumn{1}{c|}{40.37}                                                    & \multicolumn{1}{c|}{35.7}                                                     & 38.04                                                    \\ \hline
\multicolumn{1}{|c|}{\begin{tabular}[c]{@{}c@{}}Baseline + \\ RmvAll\end{tabular}}  & \multicolumn{1}{c|}{40.06}                                                    & \multicolumn{1}{c|}{38.82}                                                    & \multicolumn{1}{c|}{39.44}                                                    & \multicolumn{1}{c|}{41.01}                                                    & \multicolumn{1}{c|}{36.34}                                                    & 38.67                                                    \\ \hline
\multicolumn{1}{|c|}{\begin{tabular}[c]{@{}c@{}}Baseline +\\ Ensemble\end{tabular}}       & \multicolumn{1}{c|}{40.38}                                                    & \multicolumn{1}{c|}{39.15}                                                    & \multicolumn{1}{c|}{39.76}                                                    & \multicolumn{1}{c|}{43.17}                                                    & \multicolumn{1}{c|}{38.80}                                                    & 40.98                                                    \\ \hline
\multicolumn{1}{|c|}{\textbf{\begin{tabular}[c]{@{}c@{}}Baseline + \\ Ours\end{tabular}}} & \multicolumn{1}{c|}{\textbf{\begin{tabular}[c]{@{}c@{}}42.73 \\ \small{(+3.33)}\end{tabular}}} & \multicolumn{1}{c|}{\textbf{\begin{tabular}[c]{@{}c@{}}40.61 \\ \small{(+1.70)}\end{tabular}}} & \multicolumn{1}{c|}{\textbf{\begin{tabular}[c]{@{}c@{}}41.67 \\ \small{(+2.52)}\end{tabular}}} & \multicolumn{1}{c|}{\textbf{\begin{tabular}[c]{@{}c@{}}45.36 \\ \small{(+4.89)}\end{tabular}}} & \multicolumn{1}{c|}{\textbf{\begin{tabular}[c]{@{}c@{}}37.80 \\ \small{(+1.20)}\end{tabular}}} & \textbf{\begin{tabular}[c]{@{}c@{}}41.58 \\ \small{(+3.04)}\end{tabular}} \\ \hline
\end{tabular}
}
\end{center}
\caption{Baseline comparison on text-video \\ retrieval for Epic-Kitchens-100.}
\label{tab:result_epic_1}
}
\hspace{0.3cm}
\parbox{.35\linewidth}{
\begin{center}
\resizebox{4cm}{!}{%
\begin{tabular}{|ccc|}
\hline
\multicolumn{1}{|c|}{\textbf{Method}} & \multicolumn{1}{c|}{\textbf{nDCG (AVG)}} & \textbf{mAP (AVG)} \\ \hline
\multicolumn{3}{|c|}{\textbf{Epic-Kitchens-100}}                                                      \\ \hline
\multicolumn{1}{|c|}{RAN}             & \multicolumn{1}{c|}{41.06}               & 39.46              \\ \hline
\multicolumn{1}{|c|}{\textit{\begin{tabular}[c]{@{}c@{}}RAN +\\ Ours\end{tabular}}} &
  \multicolumn{1}{c|}{\textit{\begin{tabular}[c]{@{}c@{}}41.84 \\ (+0.78)\end{tabular}}} &
  \textit{\begin{tabular}[c]{@{}c@{}}41.24 \\ (+1.78)\end{tabular}} \\ \hline
\multicolumn{1}{|c|}{RANP}            & \multicolumn{1}{c|}{43.14}               & 43.77              \\ \hline
\multicolumn{1}{|c|}{\textit{\begin{tabular}[c]{@{}c@{}}RANP + \\ Ours\end{tabular}}} &
  \multicolumn{1}{c|}{\textit{\begin{tabular}[c]{@{}c@{}}43.80\\ (+0.66)\end{tabular}}} &
  \textit{\begin{tabular}[c]{@{}c@{}}44.12 \\ (+0.35)\end{tabular}} \\ \hline
\end{tabular}%
}
\end{center}
\caption{SOTA comparison \\on text-video retrieval for \\Epic-Kitchens-100.}
\label{tab:result_epic_2}
}
\end{center}
\end{table}

\vspace{-0.57cm}
\begin{table}[!htb]
\begin{center}
\resizebox{10cm}{!}{%
\begin{tabular}{|cccccccccc|}
\hline
\multicolumn{1}{|c|}{\multirow{2}{*}{\textbf{Method}}} &
  \multicolumn{6}{c|}{\textbf{Recall (T2V)}} &
  \multicolumn{3}{c|}{\textbf{nDCG}} \\ \cline{2-10} 
\multicolumn{1}{|c|}{} &
  \multicolumn{1}{c|}{\textbf{R@1↑}} &
  \multicolumn{1}{c|}{\textbf{R@5↑}} &
  \multicolumn{1}{c|}{\textbf{R@10↑}} &
  \multicolumn{1}{c|}{\textbf{MedR↓}} &
  \multicolumn{1}{c|}{\textbf{MnR↓}} &
  \multicolumn{1}{c|}{\textbf{Rsum↑}} &
  \multicolumn{1}{c|}{\textbf{V2T↑}} &
  \multicolumn{1}{c|}{\textbf{T2V↑}} &
  \textbf{AVG↑} \\ \hline
\multicolumn{10}{|c|}{\textbf{YouCook2}} \\ \hline
\multicolumn{1}{|c|}{Baseline} &
  \multicolumn{1}{c|}{13.17} &
  \multicolumn{1}{c|}{36.31} &
  \multicolumn{1}{c|}{50.74} &
  \multicolumn{1}{c|}{10} &
  \multicolumn{1}{c|}{66.47} &
  \multicolumn{1}{c|}{100.23} &
  \multicolumn{1}{c|}{49.42} &
  \multicolumn{1}{c|}{49.70} &
  49.56 \\ \hline
\multicolumn{1}{|c|}{\textbf{\begin{tabular}[c]{@{}c@{}}Baseline +\\ Ours\end{tabular}}} &
  \multicolumn{1}{c|}{\textbf{\begin{tabular}[c]{@{}c@{}}14.60\\ \small{(+1.43)}\end{tabular}}} &
  \multicolumn{1}{c|}{\textbf{\begin{tabular}[c]{@{}c@{}}37.80\\ \small{(+1.49)}\end{tabular}}} &
  \multicolumn{1}{c|}{\textbf{\begin{tabular}[c]{@{}c@{}}51.58\\ \small{(+0.84)}\end{tabular}}} &
  \multicolumn{1}{c|}{10} &
  \multicolumn{1}{c|}{\textbf{\begin{tabular}[c]{@{}c@{}}63.18\\ \small{(-3.29)}\end{tabular}}} &
  \multicolumn{1}{c|}{\textbf{\begin{tabular}[c]{@{}c@{}}103.98 \\ \small{(+3.75)}\end{tabular}}} &
  \multicolumn{1}{c|}{\textbf{\begin{tabular}[c]{@{}c@{}}51.92 \\ \small{(+2.50)}\end{tabular}}} &
  \multicolumn{1}{c|}{\textbf{\begin{tabular}[c]{@{}c@{}}51.39 \\ \small{(+1.69)}\end{tabular}}} &
  \textbf{\begin{tabular}[c]{@{}c@{}}51.65 \\ \small{(+2.09)}\end{tabular}} \\ \hline \hline
\multicolumn{1}{|c|}{RAN} &
  \multicolumn{1}{c|}{13.29} &
  \multicolumn{1}{c|}{36.37} &
  \multicolumn{1}{c|}{50.40} &
  \multicolumn{1}{c|}{10} &
  \multicolumn{1}{c|}{64.85} &
  \multicolumn{1}{c|}{100.06} &
  \multicolumn{1}{c|}{50.17} &
  \multicolumn{1}{c|}{50.35} &
  50.26 \\ \hline 
\multicolumn{1}{|c|}{\textit{\begin{tabular}[c]{@{}c@{}}RAN \\ + Ours\end{tabular}}} &
  \multicolumn{1}{c|}{\textit{\begin{tabular}[c]{@{}c@{}}14.92 \\ \small{(+1.63)}\end{tabular}}} &
  \multicolumn{1}{c|}{\textit{\begin{tabular}[c]{@{}c@{}}37.37 \\ \small{(+1.00)}\end{tabular}}} &
  \multicolumn{1}{c|}{\textit{\begin{tabular}[c]{@{}c@{}}50.86 \\ \small{(+0.46)}\end{tabular}}} &
  \multicolumn{1}{c|}{10} &
  \multicolumn{1}{c|}{\textit{\begin{tabular}[c]{@{}c@{}}63.78\\ \small{(-1.07)}\end{tabular}}} &
  \multicolumn{1}{c|}{\textit{\begin{tabular}[c]{@{}c@{}}103.15 \\ \small{(+3.09)}\end{tabular}}} &
  \multicolumn{1}{c|}{\textit{\begin{tabular}[c]{@{}c@{}}50.97 \\ \small{(+0.80)}\end{tabular}}} &
  \multicolumn{1}{c|}{\textit{\begin{tabular}[c]{@{}c@{}}51.25 \\ \small{(+0.90)}\end{tabular}}} &
  \textit{\begin{tabular}[c]{@{}c@{}}51.11 \\ \small{(+0.85)}\end{tabular}} \\ \hline
\multicolumn{1}{|c|}{RANP} &
  \multicolumn{1}{c|}{13.63} &
  \multicolumn{1}{c|}{35.65} &
  \multicolumn{1}{c|}{50.32} &
  \multicolumn{1}{c|}{10} &
  \multicolumn{1}{c|}{64.34} &
  \multicolumn{1}{c|}{99.60} &
  \multicolumn{1}{c|}{50.49} &
  \multicolumn{1}{c|}{50.19} &
  50.34 \\ \hline
\multicolumn{1}{|c|}{\textit{\begin{tabular}[c]{@{}c@{}}RANP + \\ Ours\end{tabular}}} &
  \multicolumn{1}{c|}{\textit{\begin{tabular}[c]{@{}c@{}}15.23 \\ \small{(+1.60)}\end{tabular}}} &
  \multicolumn{1}{c|}{\textit{\begin{tabular}[c]{@{}c@{}}37.60 \\ \small{(+1.95)}\end{tabular}}} &
  \multicolumn{1}{c|}{\textit{\begin{tabular}[c]{@{}c@{}}51.58 \\ \small{(+1.26)}\end{tabular}}} &
  \multicolumn{1}{c|}{10} &
  \multicolumn{1}{c|}{\textit{\begin{tabular}[c]{@{}c@{}}61.34\\ \small{(-3.00)}\end{tabular}}} &
  \multicolumn{1}{c|}{\textit{\begin{tabular}[c]{@{}c@{}}104.41 \\ \small{(+4.81)}\end{tabular}}} &
  \multicolumn{1}{c|}{\textit{\begin{tabular}[c]{@{}c@{}}51.53 \\ \small{(+1.08)}\end{tabular}}} &
  \multicolumn{1}{c|}{\textit{\begin{tabular}[c]{@{}c@{}}51.05 \\ \small{(+0.86)}\end{tabular}}} &
  \textit{\begin{tabular}[c]{@{}c@{}}51.29 \\ \small{(+0.95)}\end{tabular}} \\ \hline \hline \hline
\multicolumn{10}{|c|}{\textbf{MSR-VTT 1kA Split}} \\ \hline
\multicolumn{1}{|c|}{Baseline} &
  \multicolumn{1}{c|}{20.76} &
  \multicolumn{1}{c|}{47.29} &
  \multicolumn{1}{c|}{59.92} &
  \multicolumn{1}{c|}{6} &
  \multicolumn{1}{c|}{41.10} &
  \multicolumn{1}{c|}{127.97} &
  \multicolumn{1}{c|}{59.77} &
  \multicolumn{1}{c|}{60.84} &
  60.30 \\ \hline
\multicolumn{1}{|c|}{\textbf{\begin{tabular}[c]{@{}c@{}}Baseline +\\ Ours\end{tabular}}} &
  \multicolumn{1}{c|}{\textbf{\begin{tabular}[c]{@{}c@{}}24.64\\ \small{(+3.88)}\end{tabular}}} &
  \multicolumn{1}{c|}{\textbf{\begin{tabular}[c]{@{}c@{}}52.99\\ \small{(+5.70)}\end{tabular}}} &
  \multicolumn{1}{c|}{\textbf{\begin{tabular}[c]{@{}c@{}}66.09\\ \small{(+6.17)}\end{tabular}}} &
  \multicolumn{1}{c|}{\textbf{\begin{tabular}[c]{@{}c@{}}5\\ \small{(-1)}\end{tabular}}} &
  \multicolumn{1}{c|}{\textbf{\begin{tabular}[c]{@{}c@{}}26.26\\ \small{(-14.84)}\end{tabular}}} &
  \multicolumn{1}{c|}{\textbf{\begin{tabular}[c]{@{}c@{}}143.72\\ \small{(+15.75)}\end{tabular}}} &
  \multicolumn{1}{c|}{\textbf{\begin{tabular}[c]{@{}c@{}}62.67\\ \small{(+2.90)}\end{tabular}}} &
  \multicolumn{1}{c|}{\textbf{\begin{tabular}[c]{@{}c@{}}62.33\\ \small{(+1.49)}\end{tabular}}} &
  \textbf{\begin{tabular}[c]{@{}c@{}}62.50\\ \small{(+2.20)}\end{tabular}} \\ \hline \hline
\multicolumn{1}{|c|}{RAN} &
  \multicolumn{1}{c|}{21.08} &
  \multicolumn{1}{c|}{47.98} &
  \multicolumn{1}{c|}{60.95} &
  \multicolumn{1}{c|}{6} &
  \multicolumn{1}{c|}{42.28} &
  \multicolumn{1}{c|}{130.01} &
  \multicolumn{1}{c|}{59.49} &
  \multicolumn{1}{c|}{60.15} &
  59.82 \\ \hline
\multicolumn{1}{|c|}{\textit{\begin{tabular}[c]{@{}c@{}}RAN \\ + Ours\end{tabular}}} &
  \multicolumn{1}{c|}{\textit{\begin{tabular}[c]{@{}c@{}}24.54\\ \small{(+3.46)}\end{tabular}}} &
  \multicolumn{1}{c|}{\textit{\begin{tabular}[c]{@{}c@{}}53.50\\ \small{(+5.52)}\end{tabular}}} &
  \multicolumn{1}{c|}{\textit{\begin{tabular}[c]{@{}c@{}}66.70\\ \small{(+5.75)}\end{tabular}}} &
  \multicolumn{1}{c|}{\textit{\begin{tabular}[c]{@{}c@{}}5\\ \small{(-1)}\end{tabular}}} &
  \multicolumn{1}{c|}{\textit{\begin{tabular}[c]{@{}c@{}}26.91\\ \small{(-15.37)}\end{tabular}}} &
  \multicolumn{1}{c|}{\textit{\begin{tabular}[c]{@{}c@{}}144.74\\ \small{(+14.73)}\end{tabular}}} &
  \multicolumn{1}{c|}{\textit{\begin{tabular}[c]{@{}c@{}}60.95\\ \small{(+1.46)}\end{tabular}}} &
  \multicolumn{1}{c|}{\textit{\begin{tabular}[c]{@{}c@{}}61.86\\ \small{(+1.71)}\end{tabular}}} &
  \textit{\begin{tabular}[c]{@{}c@{}}61.41\\ \small{(+1.59)}\end{tabular}} \\ \hline
\multicolumn{1}{|c|}{RANP} &
  \multicolumn{1}{c|}{21.14} &
  \multicolumn{1}{c|}{47.72} &
  \multicolumn{1}{c|}{60.32} &
  \multicolumn{1}{c|}{6} &
  \multicolumn{1}{c|}{41.66} &
  \multicolumn{1}{c|}{129.18} &
  \multicolumn{1}{c|}{59.94} &
  \multicolumn{1}{c|}{60.55} &
  60.25 \\ \hline
\multicolumn{1}{|c|}{\textit{\begin{tabular}[c]{@{}c@{}}RANP + \\ Ours\end{tabular}}} &
  \multicolumn{1}{c|}{\textit{\begin{tabular}[c]{@{}c@{}}24.03\\ \small{(+2.89)}\end{tabular}}} &
  \multicolumn{1}{c|}{\textit{\begin{tabular}[c]{@{}c@{}}53.24\\ \small{(+5.52)}\end{tabular}}} &
  \multicolumn{1}{c|}{\textit{\begin{tabular}[c]{@{}c@{}}66.53\\ \small{(+6.21)}\end{tabular}}} &
  \multicolumn{1}{c|}{\textit{\begin{tabular}[c]{@{}c@{}}5\\ \small{(-1)}\end{tabular}}} &
  \multicolumn{1}{c|}{\textit{\begin{tabular}[c]{@{}c@{}}27.35\\ \small{(-14.31)}\end{tabular}}} &
  \multicolumn{1}{c|}{\textit{\begin{tabular}[c]{@{}c@{}}143.81\\ \small{(+14.63)}\end{tabular}}} &
  \multicolumn{1}{c|}{\textit{\begin{tabular}[c]{@{}c@{}}61.54\\ \small{(+1.60)}\end{tabular}}} &
  \multicolumn{1}{c|}{\textit{\begin{tabular}[c]{@{}c@{}}61.58\\ \small{(+1.03)}\end{tabular}}} &
  \textit{\begin{tabular}[c]{@{}c@{}}61.56\\ \small{(+1.31)}\end{tabular}} \\ \hline
\end{tabular}%
}
\end{center}
\caption{Baseline and SOTA comparison on text-video retrieval for YouCook2 and MSR-VTT. The lower, the better for MedR and MnR metrics; the higher, the better for the rest.}
\label{tab:main_result}
\end{table}


\begin{figure}[!htb]
    \begin{center}
    \resizebox{12.9cm}{!}{
    \includegraphics[width=1.0\linewidth]{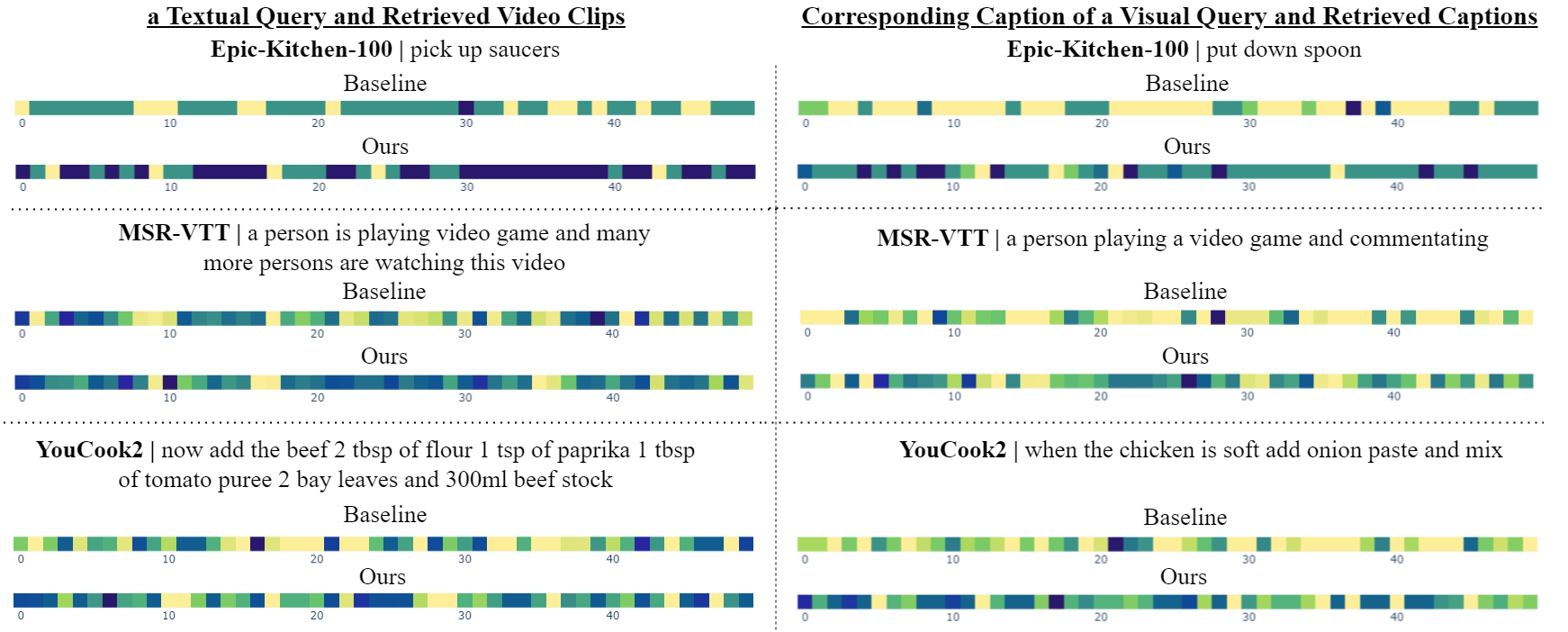}
    }
    \end{center}
    \caption{Qualitative results for text-video retrieval. The semantic relevancy, calculated based on nDCG, of the top 50 retrievals given a query from each dataset. The darker the colour, the more relevant retrievals to the query, varying from 0 to 1. While the left side is for T2V, the right side is for V2T. Best viewed in colour.}
    \label{fig:qualitative}
\end{figure}

\textbf{Qualitative Results.} Figure \ref{fig:qualitative} shows qualitative examples, proving that the bias is mitigated. We utilise the nDCG metric, knowing that we cannot examine this by using only Recall or mAP metrics due to their nature of binary similarity. Regarding text-to-video retrieval on the left side of the figure, the top retrieved video clips and the neighbour clips become more relevant than the baseline in the first example. In the second example, the top retrieved clip is already related to the query in the baseline model; however, the causal model eliminates most of the unrelated clips and provides more relevant clips in total. The third example's query is complex, but our approach still outperforms the baseline. On the right side, for video-to-text retrieval, queries are videos, and we retrieve the textual queries. However, for simplicity, we report their corresponding captions. Darker colours refer to higher relevancy. Please refer to the Supplementary Material for more analysis.

\subsection{Analysis}

\textbf{Ablation Study.} Table \ref{tab:ablation_study_1} examines three questions: \textbf{i)} \textit{How to split the dataset?} When we adjust the splits based on the frame length distribution of the dataset rather than dividing them into two equal splits, we reach a higher result. Adjusted splits have higher entropy, bringing better cooperation between splits. \textbf{ii)} \textit{Which split effects more?} When the splits are adjusted according to the frame length distribution, they share a similar score in nDCG, even though the second split has fewer video clips. Also, the first split brings a higher score in mAP/Recall, as expected. \textbf{iii)} \textit{How many splits do we need?} The more splits we have, the lower the scores we get. To have the adjusted splits when $M>2$, we first put the videos in ascending order according to the length of the frame, then divide them into two and continue to divide the remainder until we reach enough splits for the experiments. 

\begin{table}[!htb]
\begin{center}
\resizebox{11.5cm}{!}{%
\begin{tabular}{|c|cc|ccc|ccc|}
\hline
\multirow{2}{*}{\textbf{Method}} &
  \multicolumn{2}{c|}{\textbf{Epic-Kitchens-100}} &
  \multicolumn{3}{c|}{\textbf{YouCook2}} &
  \multicolumn{3}{c|}{\textbf{MSR-VTT}} \\ \cline{2-9} 
 &
  \multicolumn{1}{c|}{\textbf{nDCG (avg)↑}} &
  \textbf{mAP (avg)↑} &
  \multicolumn{1}{c|}{\textbf{nDCG (avg)↑}} &
  \multicolumn{1}{c|}{\textbf{R@10↑}} &
  \textbf{MnR↓} &
  \multicolumn{1}{c|}{\textbf{nDCG (avg)↑}} &
  \multicolumn{1}{c|}{\textbf{R@10↑}} &
  \textbf{MnR↓} \\ \hline
Baseline &
  \multicolumn{1}{c|}{39.15} &
  38.54 &
  \multicolumn{1}{c|}{49.56} &
  \multicolumn{1}{c|}{50.74} &
  66.47 &
  \multicolumn{1}{c|}{60.30} &
  \multicolumn{1}{c|}{59.92} &
  41.10 \\ \hline
\begin{tabular}[c]{@{}c@{}}Baseline + Ours\\ (Equal 2 Splits)\end{tabular} &
  \multicolumn{1}{c|}{41.19} &
  41.07 &
  \multicolumn{1}{c|}{51.01} &
  \multicolumn{1}{c|}{51.15} &
  66.22 &
  \multicolumn{1}{c|}{62.18} &
  \multicolumn{1}{c|}{66.98} &
  26.65 \\ \hline
\begin{tabular}[c]{@{}c@{}}Baseline + Ours\\ (Adjusted 2 Splits)\end{tabular} &
  \multicolumn{1}{c|}{41.67} &
  41.58 &
  \multicolumn{1}{c|}{51.65} &
  \multicolumn{1}{c|}{51.58} &
  63.18 &
  \multicolumn{1}{c|}{62.50} &
  \multicolumn{1}{c|}{66.09} &
  26.26 \\ \hline \hline
\begin{tabular}[c]{@{}c@{}}Baseline + Ours\\ (Adjusted 3 Splits)\end{tabular} &
  \multicolumn{1}{c|}{41.06} &
  39.48 &
  \multicolumn{1}{c|}{51.45} &
  \multicolumn{1}{c|}{49.28} &
  68.94 &
  \multicolumn{1}{c|}{62.48} &
  \multicolumn{1}{c|}{64.56} &
  30.54 \\ \hline
\begin{tabular}[c]{@{}c@{}}Baseline + Ours\\ (Adjusted 4 Splits)\end{tabular} &
  \multicolumn{1}{c|}{39.89} &
  37.54 &
  \multicolumn{1}{c|}{51.64} &
  \multicolumn{1}{c|}{47.11} &
  74.33 &
  \multicolumn{1}{c|}{62.23} &
  \multicolumn{1}{c|}{61.93} &
  33.74 \\ \hline \hline
\begin{tabular}[c]{@{}c@{}}First Split Only\\ (Adjusted)\end{tabular} &
  \multicolumn{1}{c|}{37.24} &
  38.59 &
  \multicolumn{1}{c|}{48.86} &
  \multicolumn{1}{c|}{48.42} &
  80.44 &
  \multicolumn{1}{c|}{61.02} &
  \multicolumn{1}{c|}{52.79} &
  50.71 \\ \hline
\begin{tabular}[c]{@{}c@{}}Second Split Only\\ (Adjusted)\end{tabular} &
  \multicolumn{1}{c|}{38.00} &
  34.07 &
  \multicolumn{1}{c|}{50.48} &
  \multicolumn{1}{c|}{36.77} &
  115.87 &
  \multicolumn{1}{c|}{59.75} &
  \multicolumn{1}{c|}{53.72} &
  50.17 \\ \hline
\end{tabular}%
}
\end{center}
\caption{Ablation study for the causal method.}
\label{tab:ablation_study_1}
\end{table}

\textbf{Computational Analysis.} Table \ref{tab:computation_cost} presents the computational cost breakdown, implemented by THOP library \cite{thop_opcounter}, on the YouCook2 dataset where the dimensions of video embedding and batch size are 1024 and 64, respectively. Considering the causal method reaches better results with two splits, we highlight two points: \textbf{i)} If it is used sequentially, there is no need for extra resources compared to the baseline method. The advantage of the causal method is that run time takes 20\% less for training. \textbf{ii)} If the splits are trained simultaneously, the run time can drop 50\% by doubling parameters and GFLOPs in return. iii) We get a similar trend in all parameters for EK-100 and MSR-VTT datasets. The only difference is that the parameters and GFLOPs are proportional to the dimension. For instance, the visual feature dimensions in EK-100 and MSR-VTT are 2048 and 3072, respectively. Thus, our approach provides a faster run time without extra resources and latency.


\begin{table}[!htb]
\begin{center}
\parbox{0.45\linewidth}{
\begin{center}
\resizebox{5.7cm}{!}{%
\begin{tabular}{c|c|c|lc|c|}
\cline{2-3} \cline{6-6}
 &
  \textbf{\begin{tabular}[c]{@{}c@{}}Text \\ Enc\end{tabular}} &
  \textbf{\begin{tabular}[c]{@{}c@{}}Video \\ Enc\end{tabular}} &
   &
   &
  \textbf{\begin{tabular}[c]{@{}c@{}}Run \\ Time (s)\end{tabular}} \\ \cline{1-3} \cline{5-6} 
\multicolumn{1}{|c|}{\begin{tabular}[c]{@{}c@{}}Parameter\\ (M)\end{tabular}} &
  10.25 &
  3.15 &
  \multicolumn{1}{l|}{} &
  Baseline &
  35 \\ \cline{1-3} \cline{5-6} 
\multicolumn{1}{|c|}{\begin{tabular}[c]{@{}c@{}}GFLOPs \\ (per clip)\end{tabular}} &
  11.93 &
  4.03 &
  \multicolumn{1}{l|}{} &
  \begin{tabular}[c]{@{}c@{}}Causal\\ (Split1/Split2)\end{tabular} &
  18 / 10 \\ \cline{1-3} \cline{5-6} 
\end{tabular}%
}
\end{center}
\caption{Computational cost breakdown on YouCook2 dataset.}
\label{tab:computation_cost}
}
\hspace{0.6cm}
\parbox{0.37\linewidth}{
\begin{center}
\resizebox{4.5cm}{!}{%
\begin{tabular}{|ccc|}
\hline
\multicolumn{1}{|c|}{\textbf{Method}} & \multicolumn{1}{c|}{\textbf{nDCG (avg)}} & \textbf{mAP (avg)} \\ \hline
\multicolumn{3}{|c|}{\textbf{Epic-Kitchens-100}}                                   \\ \hline
\multicolumn{1}{|c|}{Baseline}     & \multicolumn{1}{c|}{38.12} & 39.79 \\ \hline
\multicolumn{1}{|c|}{w/o audio}    & \multicolumn{1}{c|}{36.55} & 37.77 \\ \hline
\multicolumn{1}{|c|}{w/o spatial}  & \multicolumn{1}{c|}{36.63} & 35.63 \\ \hline
\multicolumn{1}{|c|}{w/o temporal} & \multicolumn{1}{c|}{32.56} & 32.50 \\ \hline
\end{tabular}
}
\end{center}
\caption{Comparison between spatial and temporal features.}
\label{tab:compare_spatial_temporal}
}
\end{center}
\end{table}

\textbf{Spatial vs Temporal features.} Table \ref{tab:compare_spatial_temporal} shows the importance of temporal features in the Epic-Kitchens-100 dataset such that removing them affects the result drastically. While we specifically focus on an overlooked temporal bias in this study, we note that biases in both domains should be addressed in the ideal case even though no study has achieved it yet.

\textbf{The models' effect on transformer-based models.} Noting that our approach is model-agnostic, Table \ref{tab:result_rebuttal_avg} shows the results of its implementation to transformer-based models. While limited computation resources led our model not to converge on the EgoVLP experiment, other modalities may affect our approach to the MMT experiment. Either way, we notice that the method's effect becomes limited on transformer-based models, and we share our related assertions in the Supplementary Material which could be related to the spatial biases.  

\begin{table}[!htb]
\centering
\resizebox{7.5cm}{!}{%
\begin{tabular}{|cc|cc|cc|}
\hline
\multicolumn{2}{|c|}{\textbf{Epic-Kitchens-100}}       & \multicolumn{2}{c|}{\textbf{YouCook2}}   & \multicolumn{2}{c|}{\textbf{MSR-VTT}}   \\ \hline
\multicolumn{1}{|c|}{\textbf{Method}} & \textbf{AVG} & \multicolumn{1}{c|}{\textbf{Method}} & \textbf{AVG} & \multicolumn{1}{c|}{\textbf{Method}} & \textbf{AVG} \\ \hline
\multicolumn{1}{|c|}{EgoVLP \cite{egovlp}}        & 12.53 & \multicolumn{1}{c|}{TACo \cite{Yang_Bisk_Gao_2021}}        & 53.53 & \multicolumn{1}{c|}{MMT \cite{gabeur2020mmt}}        & 63.79 \\ \hline
\multicolumn{1}{|c|}{EgoVLP + Ours} & 13.06 & \multicolumn{1}{c|}{TACo + Ours} & 54.03 & \multicolumn{1}{c|}{MMT + Ours} & 63.94 \\ \hline
\end{tabular}%
}
\vspace{0.3cm}
\caption{Comparison with transformer-based models on text-video retrieval on nDCG metric.}
\label{tab:result_rebuttal_avg}
\end{table}

\vspace{-0.5cm}
\section{Conclusion}

To the best of our knowledge, this is the first attempt to study the effect of a temporal bias caused by a frame length mismatch between training and test sets of trimmed video clips and show improvement with debiasing on the text-video retrieval task. We then discuss detailed experiments and ablation studies using our causal approach on the Epic-Kitchens-100, YouCook2 and MSR-VTT  datasets. Benchmark using the nDCG metric demonstrates that the bias has been reduced. We reckon the following limitations for future works: \textbf{i)} Long video clips may contain ambiguity, including various actions irrelevant to the annotated action. \textbf{ii)} Other temporal biases may still affect the model, e.g. the order of the actions.

\section*{Acknowledgements}

This research is supported by the Agency for Science, Technology and Research (A*STAR) under its AME Programmatic Funding Scheme (Project A18A2b0046).

\bibliography{egbib}

\newpage
\appendix

\section*{Appendix}

This paper presents the issue of frame length bias in text-video retrieval and proposes a causal intervention method to alleviate the length bias. In this Supplementary Material, we provide more details regarding the following topics. Section \textcolor{red}{A} gives more visual clues to verify the bias for each dataset. Section \textcolor{red}{B} shares more information about the baseline debiasing method. Section \textcolor{red}{C} gives more information regarding evaluation metrics, the results and the ablation study.

\section{Bias Verification}

\begin{figure}[!htb]
\begin{tabular}{cc}
\bmvaHangBox{\fbox{\includegraphics[width=5.8cm]{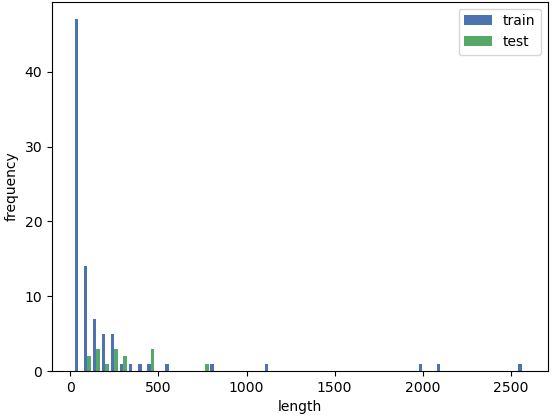}}}&
\bmvaHangBox{\fbox{\includegraphics[width=5.8cm]{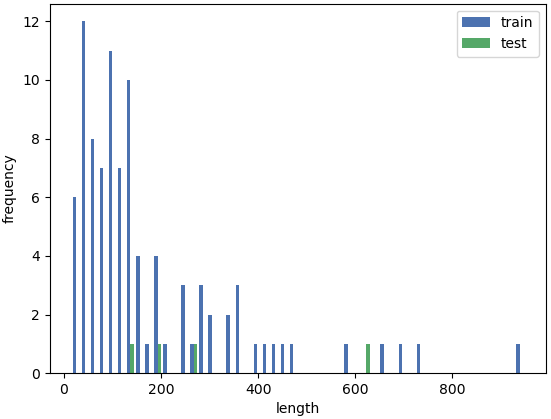}}}\\
(a) Caption: \textit{'put-down mozzarella'} & (b) Caption: \textit{'pick up rubbish'}.
\end{tabular}
\vspace{0.15cm}
\caption{The histogram shows the discrepancy between two <verb, noun> pairs (classes) in the Epic-Kitchens-100 dataset. The GT Recall is at 152nd rank and 169th rank, respectively.}
\label{fig:verify-bias-epic-only}
\end{figure}

We first provide more information that supports and verifies the bias for each dataset. For Epic-Kitchens-100, Figure \ref{fig:verify-bias-epic-only} presents the disparity for two cases. \footnote{We thank Mr. Haoxin Li for the initial discussion on verifying the bias.}

\begin{figure}[!htb]
\begin{tabular}{cc}
\bmvaHangBox{\fbox{\includegraphics[width=5.8cm]{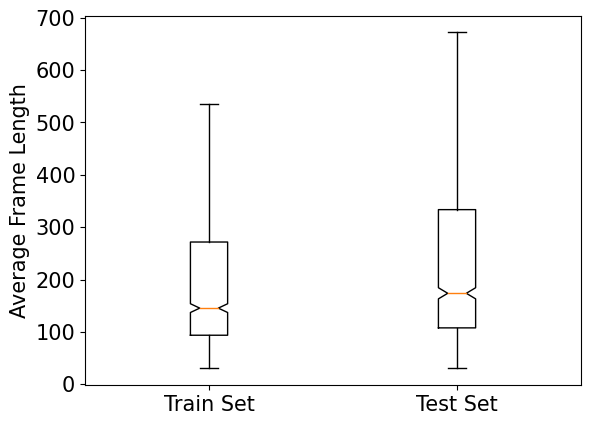}}}&
\bmvaHangBox{\fbox{\includegraphics[width=5.8cm]{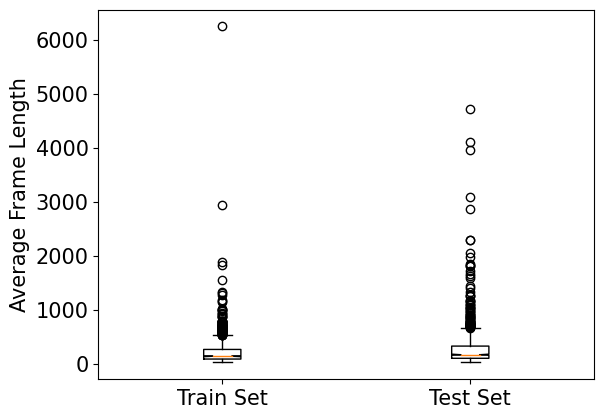}}}\\
(a) w/o Outliers & (b) w Outliers
\end{tabular}
\vspace{0.15cm}
\caption{Average frame length comparison between the training and test set in the Epic-Kitchens-100 dataset. It shows that clips in the test set are longer than the training set.}
\label{fig:verify-bias-box-epic}
\end{figure}

Figure \ref{fig:verify-bias-box-epic} depicts the discrepancy that occurs among many classes (semantic pairs of verb and noun) between the train/test set.

\begin{figure}[!htb]
    \centering
    \resizebox{10cm}{!}{
    \includegraphics{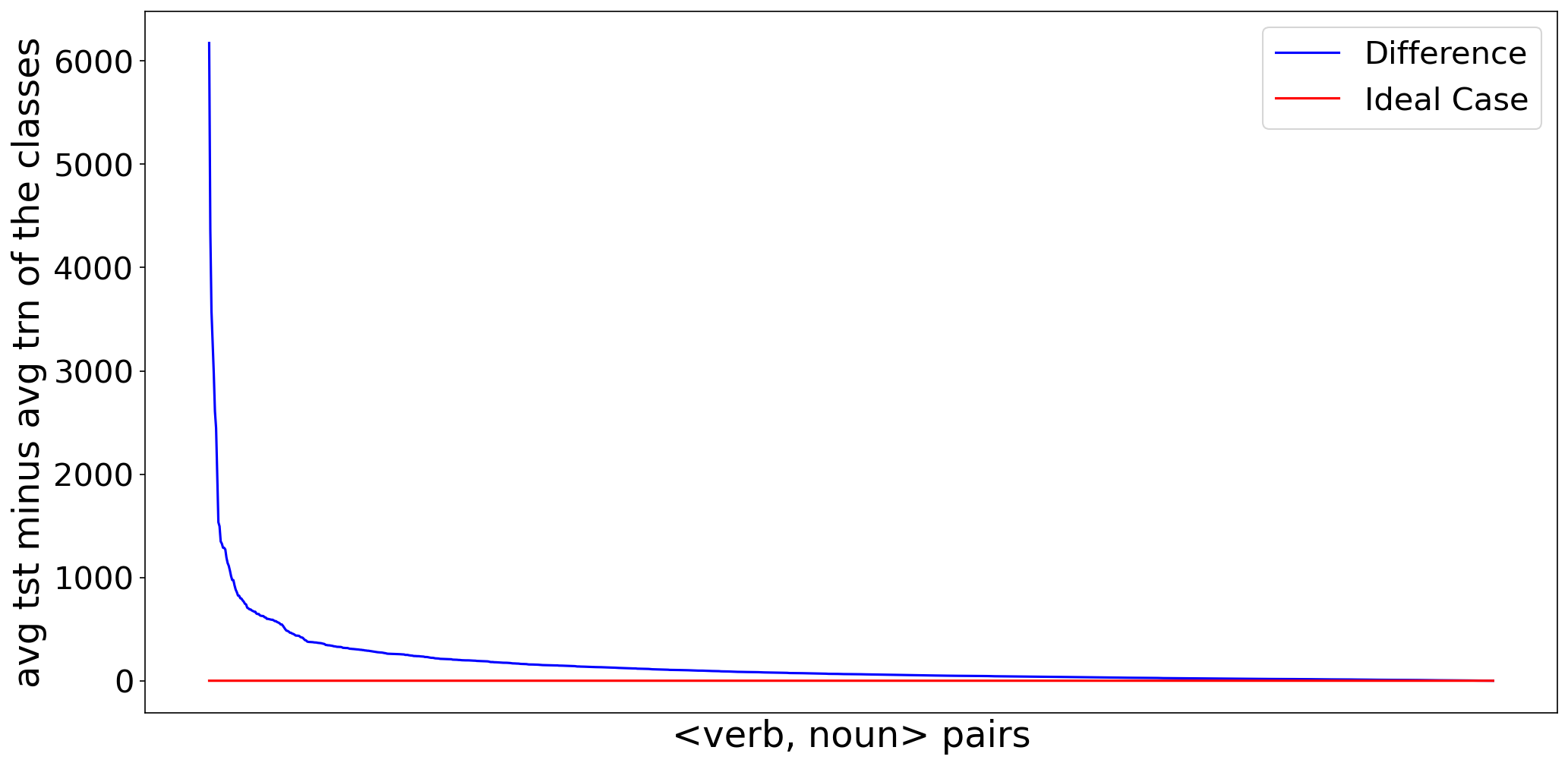}
    }
    \vspace{0.1cm}
    \caption{Long-tailed distribution among the training and test set regarding the average frame length of the classes for the Epic-Kitchens-100 dataset.}
    \label{fig:long-tail-epic}
\end{figure}

We also share another version of Figure \ref{fig:verify-bias}\textcolor{red}{a} in Figure \ref{fig:long-tail-epic}. We get the absolute value of the discrepancy among the <verb, noun> pairs in the training and test sets and sort it in descending order, showing a long-tailed distribution. While 559 videos out of 1114 have a disparity of more than 60, 212 of them have a disparity of at least 200, showing a crucial amount of biased pairs in the dataset.

\begin{figure}[!htb]
\begin{tabular}{cc}
\bmvaHangBox{\fbox{\includegraphics[width=4.95cm]{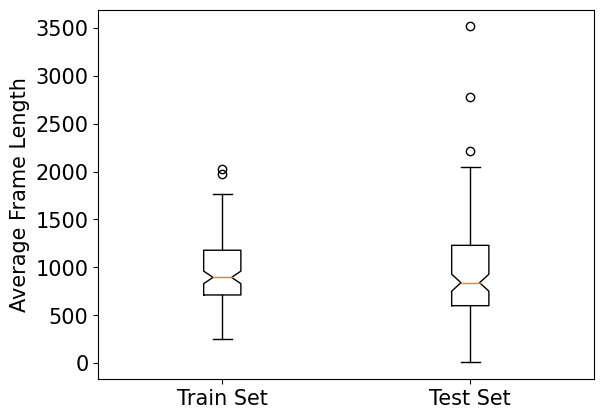}}}&
\bmvaHangBox{\fbox{\includegraphics[width=6.55cm]{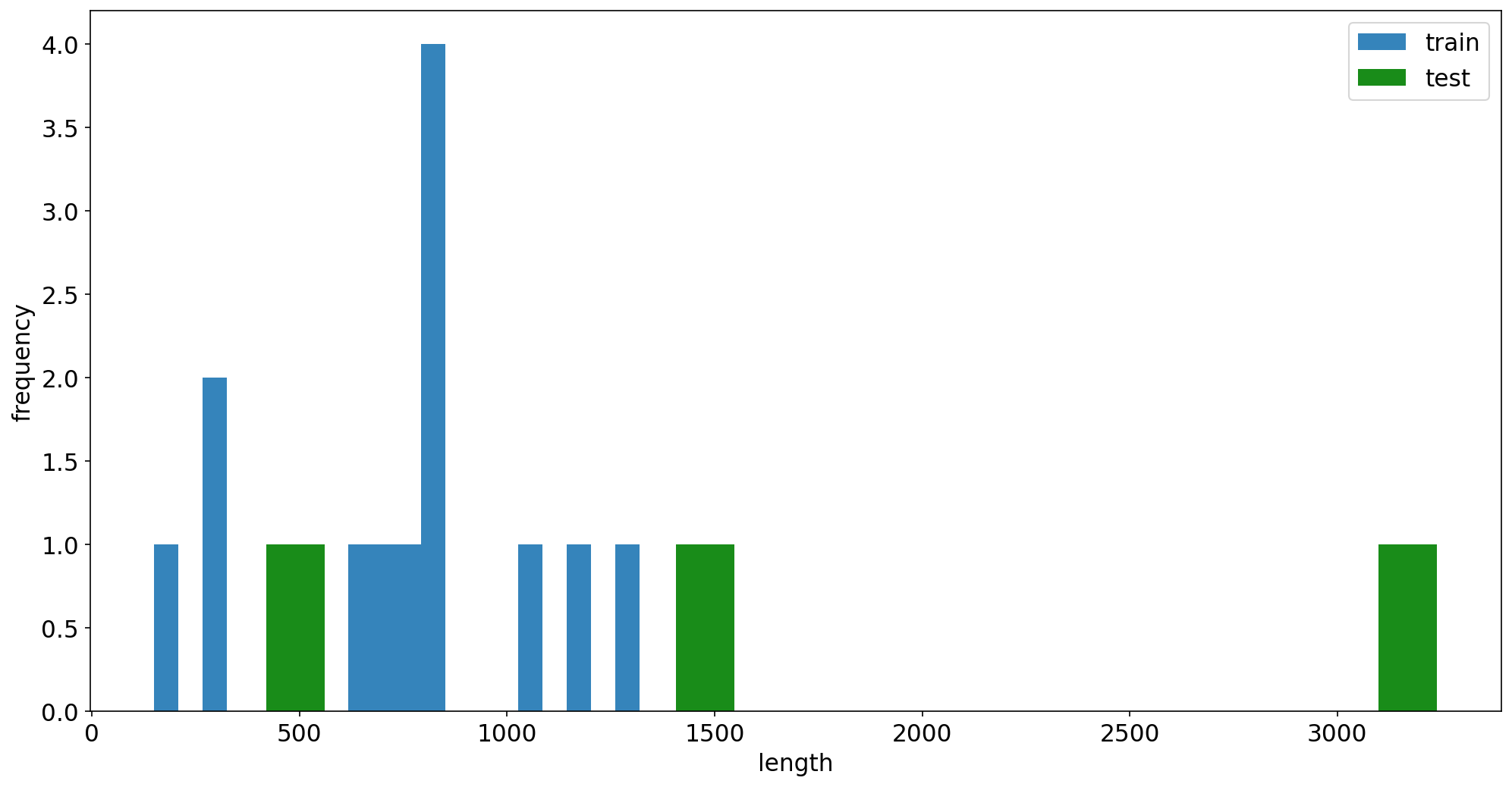}}}\\
(a) Comparison w Outliers & (b) Caption: \textit{'mix salad'}
\end{tabular}
\vspace{0.15cm}
\caption{a) The average frame length comparison between the training and test set in the YouCook2 dataset shows that clips in the test set are longer than the training set. b) An example of this disparity.}
\label{fig:verify-bias-box-yc2}
\end{figure}

\begin{figure}[!htb]
\begin{tabular}{cc}
\bmvaHangBox{\fbox{\includegraphics[width=5 cm]{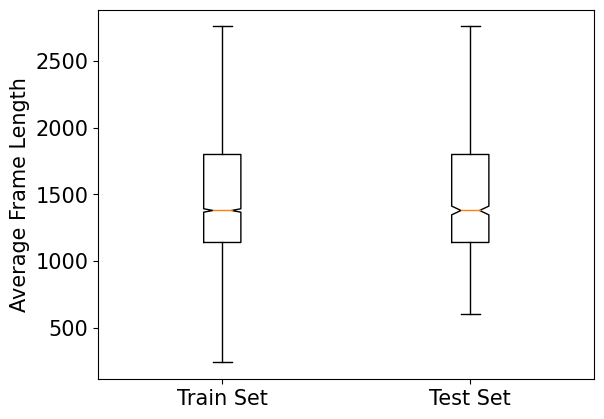}}}&
\bmvaHangBox{\fbox{\includegraphics[width=6.5cm]{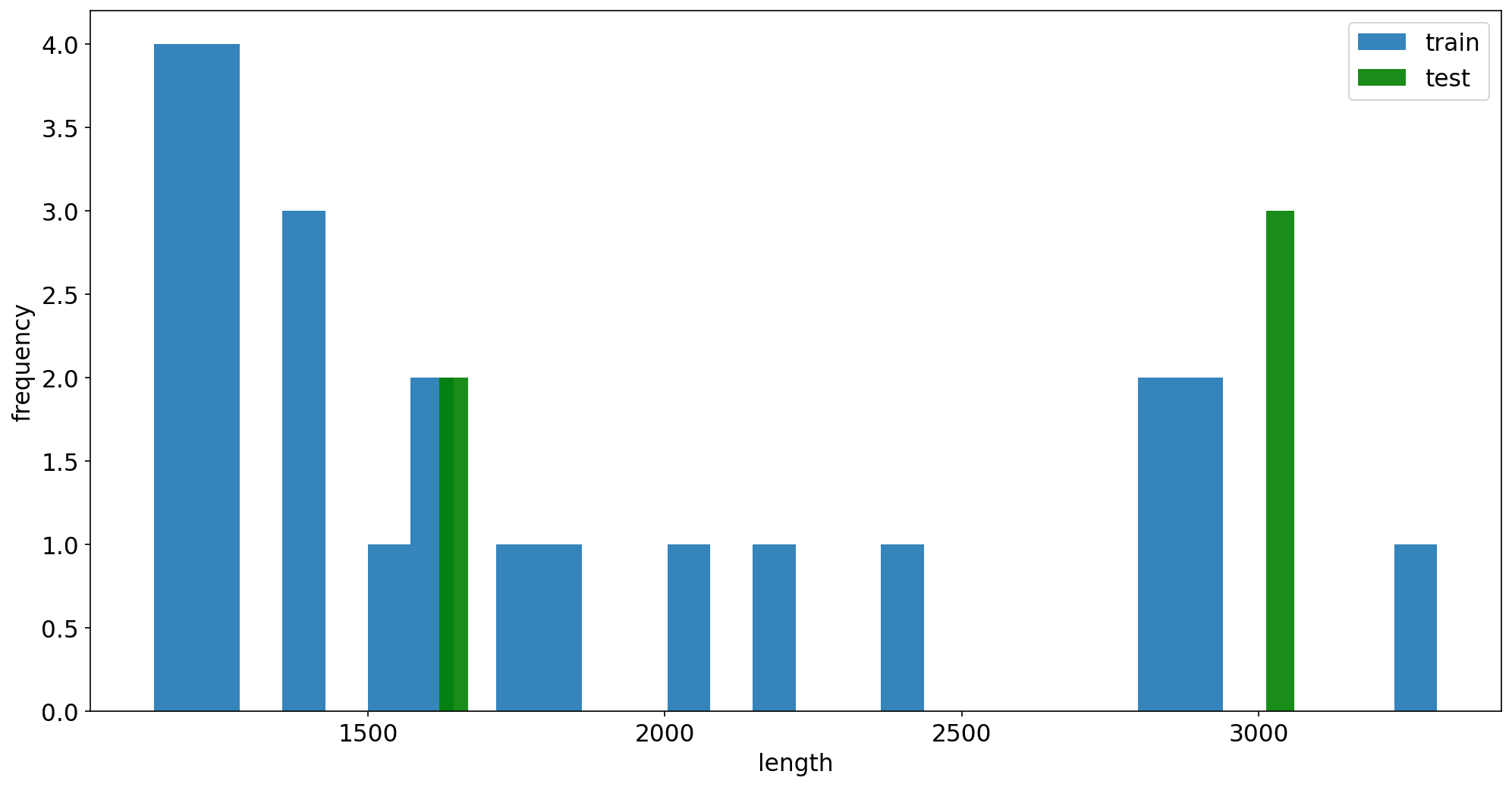}}}\\
(a) Comparison w/o Outliers & (b) Caption: \textit{'make origami'}
\end{tabular}
\vspace{0.15cm}
\caption{a) The average frame length comparison between the training and test set in the MSR-VTT dataset shows that clips in the training set are more dense than the test set. However, the discrepancy is relatively lower in the MSR-VTT compared to other datasets. b) An example of this disparity.}
\label{fig:verify-bias-box-msrvtt}
\end{figure}

The same trend can be seen for YouCook2 in Figure \ref{fig:verify-bias-box-yc2}. We note that the complex structure of the actions in YouCook2 affects the figures. For instance, 75\% of the video clips have more than one action; thus, we do not know which specific actions last how long, whereas we have this information in the Epic-Kitchens-100 dataset. Thus, we assume that each action in the same video clip lasts equally with one another. MSR-VTT also includes complex actions, as it happens in YouCook2. Besides, the MSR-VTT dataset contains coarse-grained features where the captions and visual clues define more generic actions, such as walking and playing the guitar. Thus, spatial clues such as background or scene in the dataset may bring biases. These factors limit our observation to visualise the frame length bias for the whole dataset. However, we can still see a similar trend in support of proving the bias as shown in Figure \ref{fig:verify-bias-box-msrvtt}. We argue that due to their complex and coarse-grained nature, the quantitative results are not as high as in EK-100.

\section{Baseline Debiasing Method}

We scrutinise the effect of the frame length bias on the Epic-Kitchens-100 dataset by examining the failure cases in the text-to-video retrieval, which are the ones whose retrievals are ranked over ten. However, we need to exclude some of the failure cases to eliminate any confounders. We exclude the ones \textbf{i)} if its caption includes either a tail verb or a tail noun so that we can be sure that this failure does not happen because of the long-tail distribution, \textbf{ii)} if the average frame length difference of the class between the train and test set is lower than 60, ensure that there is enough discrepancy, \textbf{iii)} if the average frame length of retrieved top 20 video clips is closer to the average frame length of the class in the test set than the average frame length of the training set. This yields 244 samples that may be affected by this bias, making the amount non-trivial. Figure \ref{fig:verify-bias-epic-only}, \ref{fig:verify-bias-box-epic}, and \ref{fig:long-tail-epic} depict the discrepancy in terms of frame length between training and testing. 

On top of the insight above, we turn captions into classes consisting of verb and noun classes by following \cite{Damen2022RESCALING} as shown in Figure \ref{fig:verify-bias} to depict this gap for the whole dataset. The difference between the average frame length of a class among training/test sets yields the discrepancy. We then visualise the discrepancy. One class includes a verb and noun describing the action. We also observe that the average frame length in the test set is higher than the one in the training set in each dataset.

These insights shape our motivation to address the bias. To this end, we first start with a baseline debiasing method. Table \ref{tab:RmvOne} shows the effect of the \textit{RmvOne} naive method on an individual sample. It indicates that if the gap between the training and test set regarding average frame length decreases, 1) top retrieved video clips increase, and 2) ground truth value at Recall improves. As shown in the Table, if we remove the shortest video clips in this class in such a way that the average frame length between training and set sets gets similar, we realise that the retrieval result gets better as well as the average frame length of the top retrieval clips converges to average frame length of the class in the test set. Besides, when we remove the longest video clips, the opposite also happens.

\begin{table}[!htb]
\centering
\resizebox{\textwidth}{!}{%
\begin{tabular}{c|cc|c|c|c|}
\cline{2-6}
\textbf{} &
  \multicolumn{2}{c|}{\textbf{Avg frame length}} &
   &
   &
   \\ \cline{1-3}
\multicolumn{1}{|c|}{\textbf{Method}} &
  \multicolumn{1}{c|}{\textbf{Train set}} &
  \textbf{Test set} &
  \multirow{-2}{*}{\textbf{\begin{tabular}[c]{@{}c@{}}\# of \\ samples\end{tabular}}} &
  \multirow{-2}{*}{\textbf{\begin{tabular}[c]{@{}c@{}}GT rank\\ recall\end{tabular}}} &
  \multirow{-2}{*}{\textbf{\begin{tabular}[c]{@{}c@{}}Avg frame length of \\ the top 10 retrieval\end{tabular}}} \\ \hline
\multicolumn{1}{|c|}{\begin{tabular}[c]{@{}c@{}}Baseline + \\ RmvOne: Remove longest clips\end{tabular}} &
  \multicolumn{1}{c|}{{\color[HTML]{333333} \textbf{64.69}}} &
  305.5 &
  43 &
  {\color[HTML]{3166FF} 261} &
  {\color[HTML]{CB0000} 131.2} \\ \hline
\multicolumn{1}{|c|}{Baseline} &
  \multicolumn{1}{c|}{{\color[HTML]{333333} \textbf{177.11}}} &
  305.5 &
  95 &
  {\color[HTML]{3166FF} 169} &
  {\color[HTML]{CB0000} 147.4} \\ \hline
\multicolumn{1}{|c|}{\begin{tabular}[c]{@{}c@{}}Baseline + \\ RmvOne: Remove shortest clips\end{tabular}} &
  \multicolumn{1}{c|}{{\color[HTML]{333333} \textbf{302.32}}} &
  305.5 &
  43 &
  {\color[HTML]{3166FF} 14} &
  {\color[HTML]{CB0000} 244.5} \\ \hline
\end{tabular}%
}
\vspace{0.3cm}
\caption{Caption: ‘pick up rubbish’ for \textit{RmvOne} naive method. While the red colour is directly proportional to the average frame length in the train set, the blue is inversely proportional.}
\label{tab:RmvOne}
\end{table}

Since it is impractical to see the effect of \textit{RmvOne} on many samples, we share how \textit{RmvAll} naive method affects the individual samples. After applying \textit{RmvAll}, we choose 60 captions randomly out of 244 suspected samples described above. We realise that 41 out of 60 samples get affected positively, while 19 out of them get negatively. When we compare the total difference between the negative and positive sides, we see a higher increase on the positive side, namely 13,321 versus 2,274. This observation double confirms the bias. We see the same trend for the other datasets as well.

Lastly, we implement our causal approach by using backdoor adjustment. It is based on the law of iterated expectations. According to this statement, the expected value of a random variable can be obtained by summing the expected values of that random variable when it is conditioned on a second random variable. To perform better in applying this method, we split the dataset into subsets so that the discrepancy in the subsets would be as much lower as possible. On the contrary, we may divide the dataset into equal splits through which various subsets could still contain high discrepancies. We follow the Algorithm \textcolor{red}{2} when the training set includes more short video clips than long ones.

\begin{algorithm}
\scriptsize
\caption{Divide the dataset into M subsets in an adjusted way:}
\begin{algorithmic}[1]
\State $V$ $\gets$ videos in ascending order based on the frame length
\State $M$ $\gets$ number of splits
\State $i$ = 1
\State 0 < $th$ < 1 \Comment{Threshold to decide the amount for the last two splits}
\While{$M \geq 1$}
\State $M_i$ = $V$[:len($V$)/2]
\If{$M$ == 2} 
    \State $M_i$ = $V$[:th*len($V$)] 
    \State $M_(i+1)$ = $V$[th*len($V$):]
    \State break; \Comment{The algorithm always finishes here}
\EndIf
\State $i$+=1
\State $M$-=1
\State $V$ = $V$[len($V$)/2:]
\EndWhile
\end{algorithmic}
\end{algorithm}

\section{Results}

\subsection{Implementation Details}

By following the algorithm above, Epic-Kitchens-100's split includes 50,282 video clips, while the second has 16,935 video clips when $M$ is selected as two. In YouCook2, the first split includes 8,000 video clips, while the second has 2,337 of them. In MSR-VTT, the first split includes 5,950, while the second has 3,050 video clips.

Another thing is that our causal approach is faster than the baseline approach. One epoch takes 305 seconds in the baseline method on the Epic-Kitchens-100 dataset. However, it takes 244 seconds in our method, while the first split and the second split take 175 seconds and 69 seconds, respectively. We use one NVIDIA RTX 2080 Ti for our experiments. In terms of sampling the positive and negative pairs, we follow the baseline study.

We assign only one relevant video to the query for text-to-video retrieval tasks using conventional evaluation metrics, such as Recall and mAP. We assign all the other videos completely irrelevant, even if there could be somewhat relevant videos. To this end, we use the nDCG metric by calculating a relevancy matrix between the captions. This calculation is done via Formula \ref{relevancy}, setting non-binary relevancy. For instance, we assume that captions are 'cut tomato', 'cut chicken' and 'take plate'. If we have a query called 'cut tomato', we calculate $R$ as the list of [1, 0.5, 0] when we apply Formula \ to these captions. However, it would be [1, 0, 0] for the conventional evaluation metrics.

\subsection{Quantitative Results} 

Table \ref{tab:main_result_msrvtt_full_split} shows that our causal method overpasses the baseline and SOTA methods for MSR-VTT full split, as well.

\begin{table}[!htb]
\begin{center}
\resizebox{10cm}{!}{%
\begin{tabular}{|cccccccccc|}
\hline
\multicolumn{1}{|c|}{\multirow{2}{*}{\textbf{Method}}} &
  \multicolumn{6}{c|}{\textbf{Recall (T2V)}} &
  \multicolumn{3}{c|}{\textbf{nDCG}} \\ \cline{2-10} 
\multicolumn{1}{|c|}{} &
  \multicolumn{1}{c|}{\textbf{R@1↑}} &
  \multicolumn{1}{c|}{\textbf{R@5↑}} &
  \multicolumn{1}{c|}{\textbf{R@10↑}} &
  \multicolumn{1}{c|}{\textbf{MedR↓}} &
  \multicolumn{1}{c|}{\textbf{MnR↓}} &
  \multicolumn{1}{c|}{\textbf{Rsum↑}} &
  \multicolumn{1}{c|}{\textbf{V2T↑}} &
  \multicolumn{1}{c|}{\textbf{T2V↑}} &
  \textbf{AVG↑} \\ \hline
\multicolumn{10}{|c|}{\textbf{MSR-VTT Full Split}} \\ \hline
\multicolumn{1}{|c|}{Baseline} &
  \multicolumn{1}{c|}{9.39} &
  \multicolumn{1}{c|}{27.60} &
  \multicolumn{1}{c|}{39.01} &
  \multicolumn{1}{c|}{20} &
  \multicolumn{1}{c|}{138.74} &
  \multicolumn{1}{c|}{76} &
  \multicolumn{1}{c|}{26.31} &
  \multicolumn{1}{c|}{24.87} &
  25.59 \\ \hline
\multicolumn{1}{|c|}{\textbf{\begin{tabular}[c]{@{}c@{}}Baseline +\\ Ours\end{tabular}}} &
  \multicolumn{1}{c|}{\textbf{\begin{tabular}[c]{@{}c@{}}10.42\\ \small{(+1.03)}\end{tabular}}} &
  \multicolumn{1}{c|}{\textbf{\begin{tabular}[c]{@{}c@{}}29.51\\ \small{(+1.91)}\end{tabular}}} &
  \multicolumn{1}{c|}{\textbf{\begin{tabular}[c]{@{}c@{}}41.68\\ \small{(+2.67)}\end{tabular}}} &
  \multicolumn{1}{c|}{\textbf{\begin{tabular}[c]{@{}c@{}}16\\ \small{(-4)}\end{tabular}}} &
  \multicolumn{1}{c|}{\textbf{\begin{tabular}[c]{@{}c@{}}84.28\\ \small{(-54.46)}\end{tabular}}} &
  \multicolumn{1}{c|}{\textbf{\begin{tabular}[c]{@{}c@{}}81.61\\ \small{(+5.61)}\end{tabular}}} &
  \multicolumn{1}{c|}{\textbf{\begin{tabular}[c]{@{}c@{}}28.27\\ \small{(+1.96)}\end{tabular}}} &
  \multicolumn{1}{c|}{\textbf{\begin{tabular}[c]{@{}c@{}}26.35\\ \small{(+1.48)}\end{tabular}}} &
  \textbf{\begin{tabular}[c]{@{}c@{}}27.31\\ \small{(+1.72)}\end{tabular}} \\ \hline \hline
\multicolumn{1}{|c|}{RAN} &
  \multicolumn{1}{c|}{9.80} &
  \multicolumn{1}{c|}{27.20} &
  \multicolumn{1}{c|}{38.28} &
  \multicolumn{1}{c|}{20} &
  \multicolumn{1}{c|}{133.53} &
  \multicolumn{1}{c|}{75.28} &
  \multicolumn{1}{c|}{26.88} &
  \multicolumn{1}{c|}{25.73} &
  26.31 \\ \hline
\multicolumn{1}{|c|}{\textit{\begin{tabular}[c]{@{}c@{}}RAN \\ + Ours\end{tabular}}} &
  \multicolumn{1}{c|}{\textit{\begin{tabular}[c]{@{}c@{}}10.90\\ \small{(+1.10)}\end{tabular}}} &
  \multicolumn{1}{c|}{\textit{\begin{tabular}[c]{@{}c@{}}30.11\\ \small{(+2.91)}\end{tabular}}} &
  \multicolumn{1}{c|}{\textit{\begin{tabular}[c]{@{}c@{}}42.20\\ \small{(+3.92)}\end{tabular}}} &
  \multicolumn{1}{c|}{\textit{\begin{tabular}[c]{@{}c@{}}16\\ \small{(-4)}\end{tabular}}} &
  \multicolumn{1}{c|}{\textit{\begin{tabular}[c]{@{}c@{}}84.68\\ \small{(-48.85)}\end{tabular}}} &
  \multicolumn{1}{c|}{\textit{\begin{tabular}[c]{@{}c@{}}83.21\\ \small{(+7.93)}\end{tabular}}} &
  \multicolumn{1}{c|}{\textit{\begin{tabular}[c]{@{}c@{}}28.73\\ \small{(+1.85)}\end{tabular}}} &
  \multicolumn{1}{c|}{\textit{\begin{tabular}[c]{@{}c@{}}26.61\\ \small{(+0.88)}\end{tabular}}} &
  \textit{\begin{tabular}[c]{@{}c@{}}27.67\\ \small{(+1.36)}\end{tabular}} \\ \hline
\multicolumn{1}{|c|}{RANP} &
  \multicolumn{1}{c|}{9.83} &
  \multicolumn{1}{c|}{27.74} &
  \multicolumn{1}{c|}{39} &
  \multicolumn{1}{c|}{20} &
  \multicolumn{1}{c|}{134.82} &
  \multicolumn{1}{c|}{76.57} &
  \multicolumn{1}{c|}{26.94} &
  \multicolumn{1}{c|}{25.37} &
  26.16 \\ \hline
\multicolumn{1}{|c|}{\textit{\begin{tabular}[c]{@{}c@{}}RANP + \\ Ours\end{tabular}}} &
  \multicolumn{1}{c|}{\textit{\begin{tabular}[c]{@{}c@{}}10.69\\ \small{(+0.86)}\end{tabular}}} &
  \multicolumn{1}{c|}{\textit{\begin{tabular}[c]{@{}c@{}}30.18\\ \small{(+2.44)}\end{tabular}}} &
  \multicolumn{1}{c|}{\textit{\begin{tabular}[c]{@{}c@{}}42.50\\ \small{(+3.50)}\end{tabular}}} &
  \multicolumn{1}{c|}{\textit{\begin{tabular}[c]{@{}c@{}}15\\ \small{(-5)}\end{tabular}}} &
  \multicolumn{1}{c|}{\textit{\begin{tabular}[c]{@{}c@{}}87.06\\ \small{(-47.76)}\end{tabular}}} &
  \multicolumn{1}{c|}{\textit{\begin{tabular}[c]{@{}c@{}}83.37\\ \small{(+6.80)}\end{tabular}}} &
  \multicolumn{1}{c|}{\textit{\begin{tabular}[c]{@{}c@{}}28.38\\ \small{(+1.44)}\end{tabular}}} &
  \multicolumn{1}{c|}{\textit{\begin{tabular}[c]{@{}c@{}}26.36\\ \small{(+0.99)}\end{tabular}}} &
  \textit{\begin{tabular}[c]{@{}c@{}}27.37\\ \small{(+1.21)}\end{tabular}} \\ \hline
\end{tabular}%
}
\end{center}
\caption{Baseline and SOTA comparison on text-video retrieval for MSR-VTT full split. \\The lower, the better for MedR and MnR metrics; the higher, the better for the rest.}
\label{tab:main_result_msrvtt_full_split}
\end{table}

\textbf{Regarding the models’ effect on transformer-based models.}
Due to our limited computation resources, we are unable to experiment with recent transformer-based methods based on heavy backbones. Even so, we argue that it would not be a fair comparison as spatial bias would heavily affect them during the finetuning, as suggested by a recent study \cite{Xiao_Tang_Wei_Liu_Lin_2023}. Moreover, another recent paper \cite{Zhang_Yang_Qi_Qian_Xu_2023} claims that they address spatial bias by using a sampling policy similar to RAN and RANP methods. Thus, our method can address the temporal bias even after mitigating the spatial bias, which would address the impact of the bias. Nevertheless, we plan to investigate the relationship between spatial and temporal biases more.

\subsection{Qualitative Results}

\begin{figure}[!htb]
    \begin{center}
    \resizebox{12.9cm}{!}{
    \includegraphics[width=1.0\linewidth]{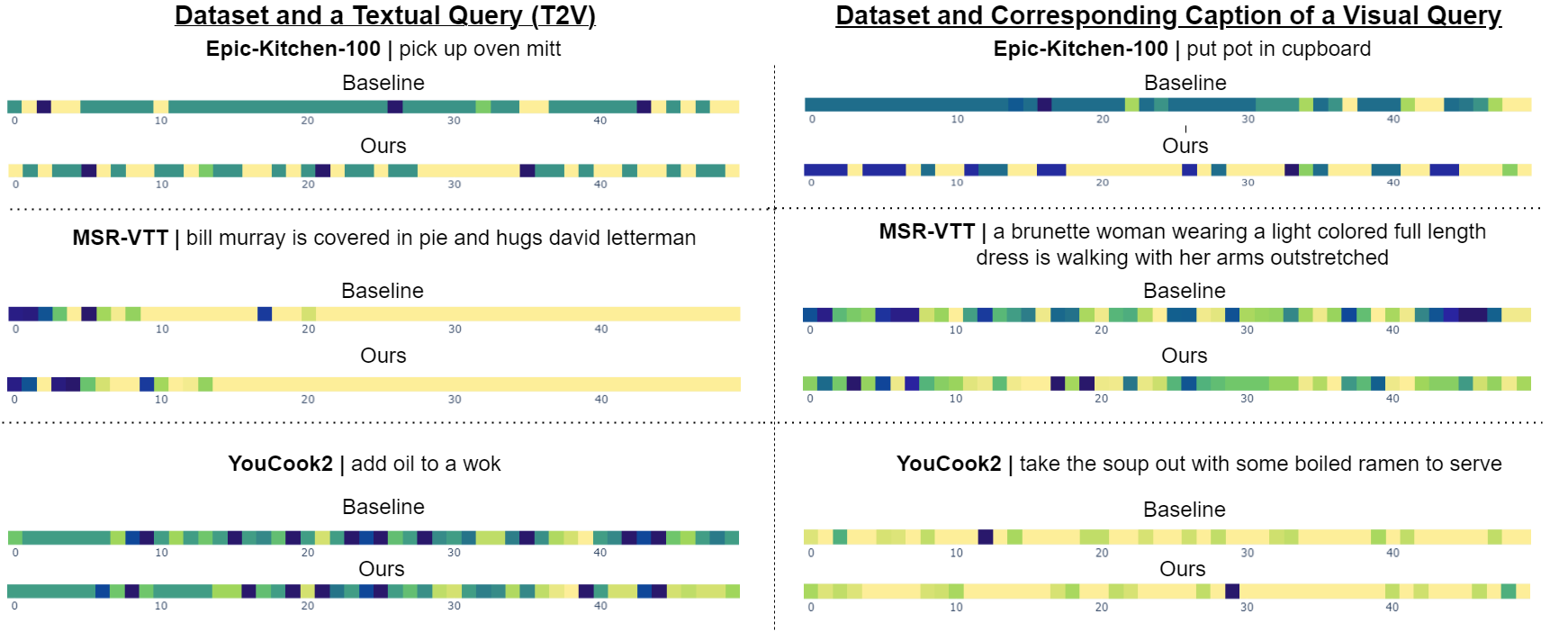}
    }
    \end{center}
    \caption{Qualitative results for text-video retrieval showing the failure cases. The semantic relevancy, calculated based on nDCG, of the top 50 retrievals given a query from each dataset. The darker the colour, the more relevant retrievals to the query, varying from 0 to 1. While the left side is for T2V and the right is for V2T. Best viewed in colour.}
    \label{fig:qualitative_failure}
\end{figure}

Figure \ref{fig:qualitative_failure} shows the failure cases on our causal method, which could occur due to various factors. For instance, considering the caption in the Epic-Kitchens-100 dataset (t2v), we see worse performance than the baseline, although the outcome is reasonable. However, we note that the caption contains a tail noun, \textit{oven mitt}. While a split may learn this feature, another split may not learn it due to fewer examples. Moreover, just summing up the similarity feature may not guarantee to preservation of the learned feature regarding various actions.

If we examine another caption in the YouCook2 dataset (t2v), we can see that the causal approach reach better result within the first ten retrieved video clips. However, the baseline method gets a higher result within fifty retrieved clips. We assume that the ambiguity of the noun, \textit{oil}, may affect the result. In other words, the difference in vector space regarding the noun could be high among the splits.

\subsection{Ablation Study}

Table \ref{tab:ablation_study_ensemble} shares the result for an ablation study between an ensemble method and our causal approach in each dataset.

\begin{table}[!htb]
\centering
\resizebox{7.5cm}{!}{%
\begin{tabular}{|c|ccc|}
\hline
\multirow{2}{*}{\textbf{Method}} & \multicolumn{1}{c|}{\textbf{Epic-Kitchens-100}} & \multicolumn{1}{c|}{\textbf{YouCook2}} & \textbf{MSR-VTT} \\ \cline{2-4} 
                    & \multicolumn{3}{c|}{\textbf{nDCG (avg)}}                        \\ \hline
Baseline            & \multicolumn{1}{c|}{39.15} & \multicolumn{1}{c|}{49.56} & 60.30 \\ \hline
Baseline + Ensemble & \multicolumn{1}{c|}{39.76} & \multicolumn{1}{c|}{49.80} & 60.60 \\ \hline
Baseline + Ours                  & \multicolumn{1}{c|}{\textbf{41.67}}        & \multicolumn{1}{c|}{\textbf{51.65}}    & \textbf{62.50}   \\ \hline
\end{tabular}%
}
\vspace{0.2cm}
\caption{Ablation study for the ensemble method.}
\label{tab:ablation_study_ensemble}
\end{table}

\textbf{A failure method.} We also try another approach; however, it does not bring any sharp increase in Precision or Recall as it gets sharply lower results in the nDCG metric. In this approach, we first train a model on the full dataset. Then, we finetune it on the data splits. We assume that finetuning works as parameter sharing, which harms the backdoor adjustment; in other words, it re-creates the link between L and V, which is shown in Figure \ref{fig:structural_causal_model}.

\textbf{Future Work.} We aim to address the same bias in related tasks such as video corpus moment retrieval, video moment retrieval \cite{ji_cvpr_binary_2023} and video localisation \cite{xiao2022boundary}. Besides, we plan to address various biases in a collaborative method.

\end{document}